\documentclass{article}

\usepackage[english]{babel}

\usepackage[letterpaper,top=2cm,bottom=2cm,left=3cm,right=3cm,marginparwidth=1.75cm]{geometry}

\usepackage{amsmath}
\usepackage{graphicx}
\usepackage{float}
\usepackage{enumitem}
\usepackage{makecell}
\usepackage{authblk}
\usepackage{cite}
\usepackage[normalem]{ulem}
\usepackage[colorlinks=true, allcolors=blue]{hyperref}

\usepackage{booktabs}
\usepackage{soul}
\usepackage{xcolor}

\usepackage{wrapfig}

\title{Deep Wave Network for Modeling Multi-Scale Physical Dynamics}

\date{}

\author[1]{Alexander I. Khrabry\thanks{Corresponding author: \href{mailto:akhrabry@princeton.edu}{akhrabry@princeton.edu}}}
\author[2]{Edward A. Startsev}
\author[3]{Andrew T. Powis}
\author[3]{Igor D. Kaganovich}

\affil[1]{Princeton University, Princeton, New Jersey 08544, USA}
\affil[2]{FusionSoft LLC, Lawrenceville, New Jersey 08648, USA}
\affil[3]{Princeton Plasma Physics Laboratory, Princeton, New Jersey 08540, USA}

\begin{document}
\maketitle

\begin{abstract}
Performance of virtually any deep learning model is strongly governed by its architectural capacity, with width and depth serving as two primary controls. Yet, model comparisons in the physical-science applications are often reported at a single chosen model size or by treating accuracy and computational cost separately, which can be misleading because different architectures exhibit different accuracy–cost scaling as width and depth vary. This issue is especially pronounced for U-Net-type encoder-decoder models, which are widely used for multiscale gas, fluid, and plasma dynamics because they efficiently model features across spatial scales. A U-Net builds a multiresolution representation through a down-sampling (encoder) path that progressively reduces spatial resolution, followed by an up-sampling (decoder) path that restores resolution for prediction. Skip connections link encoder and decoder features at the same spatial resolution, preserving fine-scale information and improving gradient flow. In practice, U-Net width (channels per resolution) is routinely tuned, while depth is commonly kept effectively fixed (a fixed number of down/up-sampling stages and a small number of convolutions per stage), limiting systematic exploration of depth as a means to improve the accuracy-cost trade-off. We address this limitation by increasing effective depth through stacking multiple encoder-decoder “waves” (U-Net modules) in series, with skip connections both within each wave and across successive waves to enable progressive cross-scale feature refinement. We term this architecture a Deep Wave Network (DW-Net). We keep the training data and learning schedule identical across models and quantify computational cost using measured GPU time. Rather than comparing isolated configurations, we train multiple width variants of each architecture and compare accuracy–GPU-time Pareto fronts. Across several published benchmarks of 2D and 3D flows, DW-Net models with a few waves improve the Pareto frontier relative to single-wave U-Net baselines, achieving higher accuracy at matched GPU time or comparable accuracy at lower GPU time, and reaching low-error thresholds with up to 3× less training time under the same training schedule.
\end{abstract}

\section{Introduction}

Accurate prediction of the dynamics of complex multi-scale physical systems is essential in many fields, including high Reynolds number fluid dynamics \cite{W98}, magnetized plasma systems \cite{H12}, weather forecasting, and atmospheric modeling \cite{Lam3}. The dynamics of these systems are characterized by the emergence of interacting structures across a wide range of spatial scales, manifesting in nonlinear phenomena such as turbulent energy cascades \cite{K61}. In such systems, the ratio between the largest and smallest relevant scales can span multiple orders of magnitude, making high-fidelity numerical simulations computationally expensive or infeasible. Moreover, resolving fast information propagation often necessitates implicit time integration schemes, further increasing computational cost. This has motivated growing interest in machine learning (ML) models for fast prediction of the systems evolution. 

In this context, ML can be used in several roles \cite{W4, L23}: (1) to accelerate traditional solvers by modeling sub-grid physics \cite{U1, K1, G3, A0, B19, L19, S1, OS20} or learning effective initial conditions \cite{Powis2025}; or (2) to fully replace solvers via learned surrogates \cite{S19, B3, A4, Z0, Lu1, W1, Khrabry2026}. Surrogate models can be further divided into two categories: (1) physics-informed neural networks (PINNs), which incorporate known governing equations into the loss function during the model training \cite{R19, K21, D4, Z1, Z3, J1, E2, li2024physics}, and purely data-driven models that rely only on observed data \cite{C4, S2, W20, G4, L3, Z4, K19}. In this work, we focus on the data-driven setting, motivated by scenarios where training data has a low temporal or spatial resolution which renders in not directly applicable to the underlying partial differential equations (PDEs). While our experiments are conducted on high-fidelity numerical solutions of known PDEs, the data was sampled at relatively large output time steps, which is a common practical setup \cite{S2} that accelerates surrogate evaluation but leads to violations of the original numerical constraints. Surrogate models are used for tasks ranging from predicting the detailed evolution of a system over a limited time horizon. which are useful, e.g., in predictive control, to making long-term predictions beyond the Lyapunov time, where accurately forecasting the system's detailed evolution becomes infeasible and statistical characteristics are needed, which is particularly useful for optimization studies.

U-Net architecture has become very popular in modeling 2D and 3D multi-scale gas, liquid, and plasma systems \cite{Z4, G2, ohana2024well}. It effectively captures multi-scale features in these turbulent systems and the interactions between them, utilizing a down-sampling path followed by an up-sampling path. At each subsequent layer in the down-sampling path, representations have 2x lower spatial resolution, which is compensated by an increasing feature map size. Skip connections between these paths at each resolution layer alleviate the loss of information and improve training, especially when it comes to retaining fine-grained details. While U-Net width (number of channels per resolution layer) is routinely tuned, its effective depth is commonly kept fixed (a fixed number of down/up-sampling stages with a small number of convolutions per stage), limiting systematic exploration of depth as a means to improve the accuracy-cost trade-off. Moreover, U-Net can exhibit temporal misalignment between encoder and decoder embeddings at the same resolution layer, as skip-connected features correspond to different effective times, particularly when predicting the temporal evolution of dynamic systems \cite{Z4}. To address this limitation, we introduce Multi-Wave Network (DW-Net), a deep learning architecture designed for efficient modeling of multi-scale physical dynamics. DW-Net builds upon the U-Net architecture by stacking multiple encoder-decoder “waves” (i.e., U-Nets), connected not only within each wave but also across waves via skip connections at matched spatial resolutions. This design enables repeated interactions across spatial scales, progressive refinement of learned dynamics, and explicit control over network depth, allowing for a more flexible and effective exploration of the depth-accuracy trade-off in complex physical system modeling.

{\bf Our contributions are:}
\begin{itemize}
    \item We present DW-Net, an efficient convolutional architecture for learning multi-scale dynamics in complex physical systems.
    \item We evaluate DW-Net on four challenging physical systems: 2D Kolmogorov turbulence, 2D Hasegawa–Wakatani (HW) plasma turbulence, buoyant smoke flow (2D and 3D), and a 2D shallow-water planetary atmosphere. DW-Net demonstrated consistent improvements over strong state-of-the-art (SOTA) baselines  achieving substantially lower prediction error and 3× faster convergence compared to best-performing baselines.
    \item For systems with longer Lyapunov times (i.e., Kolmogorov flow, buoyant smoke, and the shallow-water planetary system), we compare the predicted trajectory to the ground truth data (high-fidelity numerical solution) for exact frame-wise agreement. In contrast, for the HW plasma turbulence, which has a with very short Lyapunov, we compare statistical characteristics of the predicted trajectory to those of the ground truth, as exact frame-wise alignment is not feasible due to the system's chaotic dynamics.
\end{itemize}

{\bf The paper is organized as follows:}

Section \ref{Related} outlines various types of surrogate models used in the literature to learn multi-scale physical dynamics and introduces the U-Net architecture. Section \ref{Models} describes the different U-Net variants compared in this study, including our DW-Net. Section \ref{Syst} introduces the physical systems upon which the models are tested. Section \ref{Experiments} details the experimental setup designed to ensure a fair comparison between models and presents the results of the model comparison. Finally, Section \ref{Conclusion} summarizes the conclusions of the study.

\section{Surrogate Models for Multi-Scale Physics (Related Work)}\label{Related}

Below we summarize major architectural families used to build surrogate models in multi scale physics, highlighting their mechanisms, strengths, and limitations.

\subsection{Fourier Neural Operators (FNOs)}

Fourier Neural Operators (FNOs) \cite{P2, T3, He3, Z0, R2} learn dynamics of the system in the spectral (Fourier) space. An FNO consists of a series of spectral convolution layers, where each layer: (1) applies a Fast Fourier Transform (FFT) to map the field to the frequency domain; (2) truncates the frequency domain to a subset of Fourier modes and applies learnable complex-valued weights to advance this modes in time; and (3) uses an inverse FFT to return to the spatial domain.

The ability of FNOs to naturally capture multiscale interactions through the parameterization of Fourier modes makes them well-suited to model Tokamak plasmas \cite{li2024plasma, carey2025neural} and turbulent fluid flows \cite{luo2024fourier, liu2025spatiotemporal}. The \textbf{strengths} of this architecture include: (1) Efficient access to global receptive fields — a few spectral modes can effectively capture large-scale patterns. (2) It provides a compact basis for smooth fields and long-range correlations, where the complexity per layer remains low, scaling roughly with the FFT cost (O(NlogN) for N grid points).

Despite their intrinsic multi-scale nature and computational efficiency, FNOs have significant \textbf{limitations} that constrain their broader adoption: (1) Frequency domain filtering can act as a low-pass filter, making it harder to model sharp local features (e.g., thin filaments, sharp fronts), leading to a potential loss of fine-scale fidelity \cite{liu2025enhancing, george2022incremental, roberts2025learning, guan2023fourier}. (2) Application to non-periodic domains or complex geometries often requires windowing or alternative bases \cite{qin2025modeling}. These limitations motivate the development of architectures that preserve nonlocal coupling while explicitly modeling local interactions.

\subsection{Transformers}

Transformers for physics are inspired by vision-style transformers \cite{V17, D0} and adapted to 2D/3D domains \cite{M4, C022, C0, Z2, N3, A4}. A transformer model first partitions the physical domain into equal-sized square or cubic patches, embedding each patch as a token vector. Self-attention layers are then applied, where the attention mechanism exchanges information across all tokens, enabling the model to learn multi-scale interactions.

The strengths of this architecture lie in the self-attention mechanism, which naturally captures multi-scale behavior: any patch can attend to any other patch. The limitations of transformer models are as follows: (1) Attention cost is quadratic in the number of tokens (which is proportional to the area in 2D or volume in 3D), limiting model scalability. Physics-oriented works have proposed linear or near-linear attention \cite{L023, Cao1, Hao3, Li3}, but these lead to a trade-off in accuracy, often comparable to simpler architectures (e.g., FNO) \cite{L023}. (2) Transformers are often data-hungry \cite{abdel2022large}. (3) In periodic domains (common in physical modeling), learnable relative (periodic) \cite{S18, Wu21} or rotary \cite{Su2024} encodings add to model complexity. These trade-offs motivate a search for alternative approaches, such as convolutional architectures, which are appealing due to their simplicity, computational efficiency, and strong inductive biases.

\subsection{Convolutional Models}

Convolutional models remain a popular choice for surrogates due to simplicity, inductive biases (translation equivariance), parameter efficiency (weight sharing), and linear scaling with the grid size for fixed kernel sizes. Convolutions are particularly effective in capturing local patterns, making them well-suited for problems where spatial locality is important and large-scale data or fine-grained features need to be processed efficiently.

\subsubsection{Encode–Process–Decode Approach}

The Encode–Process–Decode approach \cite{B18, SG18, SG20} is frequently used in fluid and plasma surrogates \cite{C21, S2, K19}. In this approach, an encoder downsamples the input data to a lower-resolution latent grid with an increased number of channels. The latent representation is then processed by a processor, often implemented as a deep ResNet \cite{H16}, which operates at a fixed spatial resolution to model the time evolution of the system. Finally, a decoder upsamples the time-advanced latent representation back to the original resolution.

This approach is simple and computationally efficient. It is \textbf{effective} for modeling local interactions at the processor’s resolution. Residual connections in the ResNet facilitate training and allow for greater model depth. The primary \textbf{limitation} is that the processor operates at a single resolution level, meaning that cross-scale interactions must be learned indirectly through the encoder-decoder pathway. This can reduce model fidelity and accuracy in systems with strong multi-scale coupling, such as energy cascades or wave–eddy interactions.

To expand receptive fields at a fixed resolution, two strategies are commonly used: (1) employing dilated convolutions, and (2) using multi-scale processors that operate at multiple resolution levels, as in U-Net or HEAP \cite{Khrabry2026}.

\subsubsection{Dilated ResNets (DilResNet)}

DilResNet replaces or augments standard convolutions with dilated convolutions to enlarge the receptive field without pooling \cite{S2}. A sequence of varying dilation ratios, such as 1, 2, 4, 8, 4, 2, 1, is commonly applied. This allows the model to capture long-range dependencies while preserving the native grid resolution and fine details.

However, larger effective kernels substantially increase compute and memory requirements, \textbf{limiting} the practical applicability of the model. In practice, DilResNet can require up to an order of magnitude more computation than comparable convolutional models with similar parameter counts \cite{Z4, G2, L023}. These limitations motivate multi-resolution designs, such as U-Net or HEAP \cite{Khrabry2026}, which natively route information across scales while maintaining efficiency.

\subsubsection{U-Net and Variants}

\begin{figure}[hbt!]
  \centering
  \scalebox{1}[0.9]{\includegraphics[width=1.0\textwidth]{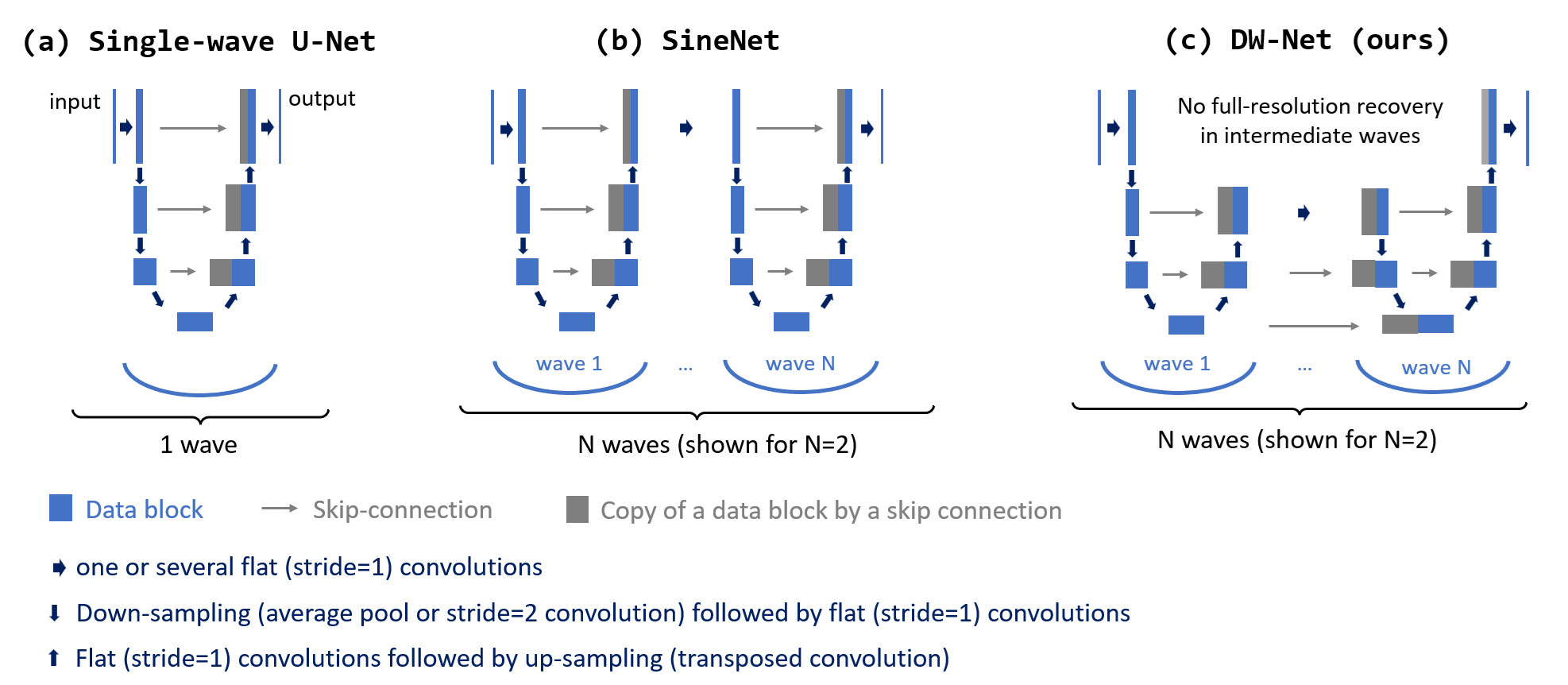}}
  \caption{Schematic of U-Net variants.  More details on the architecture design are shown in Fig. \ref{UNets}. The models are illustrated with 4 resolution levels for simplicity, but more common 5-level variants were used in this study.}
  \label{UNetsSimple}
\end{figure}

A U-Net couples a multi-scale encoder and decoder via skip connections at matching resolution levels (Fig. \ref{UNetsSimple}a), allowing features at each scale from the encoder to directly inform the decoder. In the encoder, down-sampling operations, such as average pooling or stride-2 convolutions with a 2×2 kernel, progressively reduce spatial resolution, and each down-sampling is typically followed by one or more flat (stride-1) convolutions that advance the representations in time. Symmetrically, the decoder progressively applies up-sampling operations (typically transposed convolutions with a 2×2 kernel) each followed by one or more flat convolutions to refine the representations at higher resolutions.

Skip connections are applied to copy encoder representations directly into the decoder, where they are concatenated channel-wise with the corresponding decoder features. Effectively, the processor is integrated within the encoder-decoder pair, operating across multiple resolution levels through these hierarchical interactions.

\textbf{Strengths:} Thanks to its strong multi scale inductive bias, computational efficiency, and robust training dynamics, U-Nets (and their many variants) are arguably the most widely adopted baseline for learned surrogates in multi scale physics, often achieving competitive—frequently state of the art—accuracy at a favorable accuracy–cost balance across diverse benchmarks \cite{Z4, G2, ohana2024well}. Common variants include Classic U-Net, Attention U-Net, ResUNet \cite{diakogiannis2020resunet}, and ConvNeXtU-Net \cite{ohana2024well}.

\textbf{Limitations and remedies:} (1) U-Nets perform only a single downsampling and upsampling pass, which limits the number of explicit cross-scale interactions during feature processing. (2) As a time-stepping predictor, a U-Net can exhibit temporal misalignment between encoder and decoder embeddings because skip-connected features correspond to different effective times when predicting the next step. SineNets \cite{Z4}, Fig. \ref{UNetsSimple}b, mitigate this by stacking multiple U-Nets sequentially \cite{X17, So18}, each advancing by a smaller sub-step to reduce misalignment. However, SineNet’s skip connections remain confined within each U-Net wave, and information flows across waves only through composition at the highest-resolution level \cite{Z4}. Attention-augmented variants further strengthen cross-scale interactions but at the cost of higher training complexity and optimization difficulty.

Unlike SineNet, which restricts skip connections to within each wave, our multi-wave U-Net variant, DW-Net (Fig. \ref{UNetsSimple}c), introduces skip connections both within and across successive waves at matching spatial resolutions. This design enhances hierarchical representation learning by enabling repeated interactions between feature representations at the same and different spatial scales. As a result, DW-Net supports progressive refinement of learned dynamics and provides explicit control over network depth through the number of stacked waves.

\section{Models: U-Net Variants to be Compared}\label{Models}

We selected baseline models that have demonstrated strong performance as physics surrogates on multiple systems in prior literature \cite{Z4, G2, ohana2024well}, in order to compare their performance with our model. Our focus is on U-Net variants, which consistently outperform other architectures across multiple benchmarks, including the two multi-scale systems evaluated in this paper. For instance, in 2D shallow water and buoyant smoke systems (described in Sections \ref{S-smoke} and \ref{S-SW}), U-Net variants outperformed Fourier Neural Operator (FNO) models by nearly an order of magnitude in terms of inverse error-to-computational-cost ratio \cite{Z4}. While we do not directly benchmark against FNOs and transformer-based models in this study, our evaluation indirectly reflects their performance through comparative analysis on shared systems.

\begin{figure}[hbt!]
  \centering
  \scalebox{1}[0.9]{\includegraphics[width=1.0\textwidth]{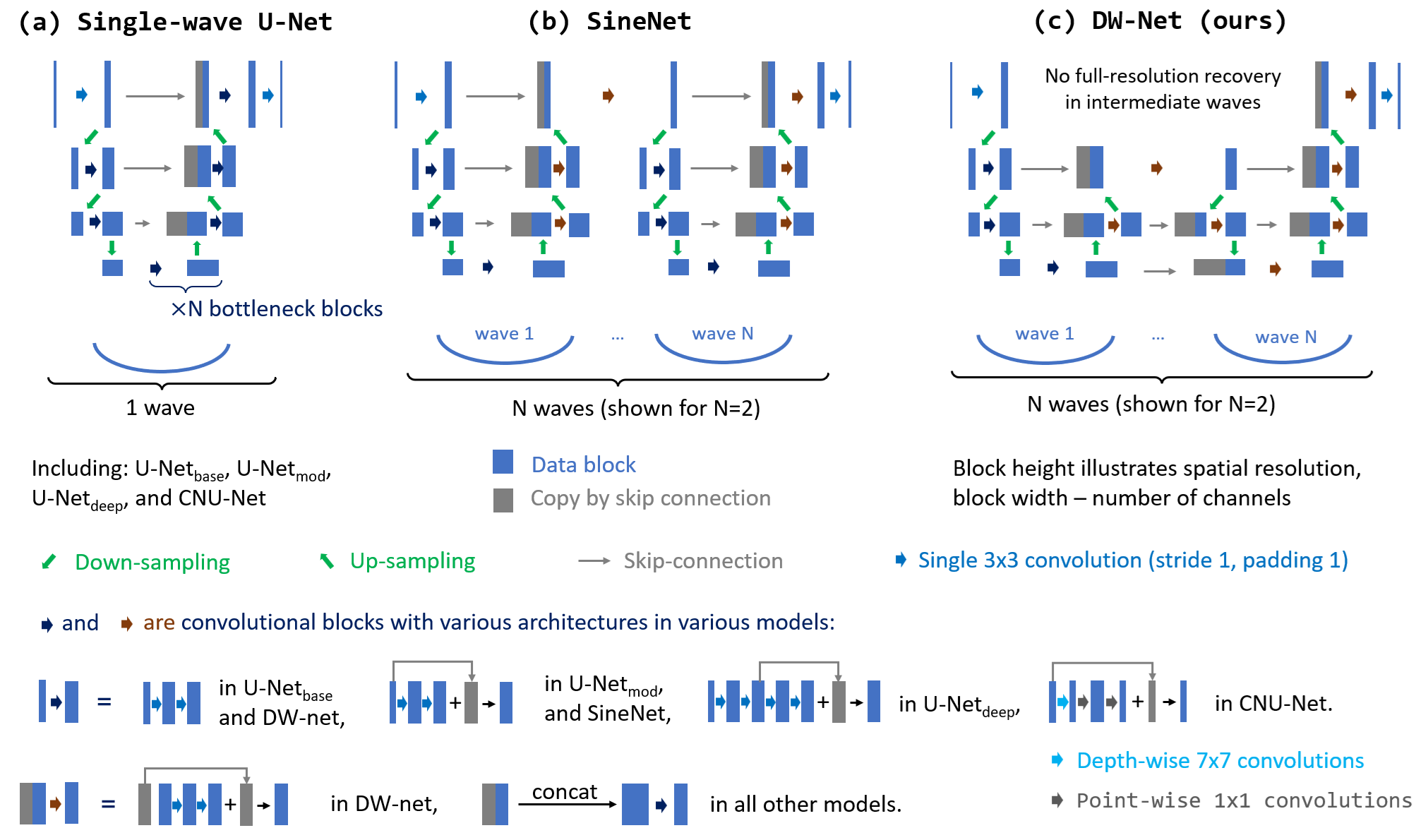}}
  \caption{Details of the U-Net variants to be compared. The models are illustrated with 4 resolution levels for simplicity (but more common 5-level variants were used in the experiments). }
  \label{UNets}
\end{figure}

\subsection{U-Net\textsubscript{base} (Base Variant)}

Our primary baseline is the U-Net architecture adapted from \cite{G2} (labeled U-Net\textsubscript{base}; Fig. \ref{UNets}a). This model closely resembles the original U-Net \cite{R15} as implemented in PDEBench \cite{M2}, with minor modifications inspired by modern variants, which include: (1) group normalization \cite{wu2018group} instead of batch normalization, (2) enabling of bias parameters in convolutional layers, (3) a reduction of bottleneck block at the lowest resolution level to match the parameter count of the original U-Net (corresponding to N=1 in the Fig. \ref{UNets}a).

Each encoder and decoder level contains a convolutional block with two 3x3 convolutions (stride 1, padding 1), a widely adopted standard in convolutional architectures \cite{Zagoruyko2016}. At the highest-resolution level, the first convolution expands the number of channels to a specified width, while the final convolution restores it to the original count. Downsampling uses 2x2 max pooling, and upsampling uses transposed convolutions.

\subsection{U-Net\textsubscript{mod} (Modernized Variant)}

The U-Net\textsubscript{mod} variant (also adapted from \cite{G2}) builds on the U-Net\textsubscript{base} variant but incorporates several enhancements from modern U-Net designs \cite{ho2020denoising, nichol2021improved, ramesh2021zero}:
\begin{itemize}
    \item Residual skip connections are used within each convolutional block, similarly to how it is done in Wide ResNet \cite{Zagoruyko2016} and ResUNet-a \cite{diakogiannis2020resunet}.
    \item Learnable downsampling via $2\times2$ convolutions (stride=2) replaced max pooling.
    \item An enhanced (deeper) bottleneck block is used (corresponding to $N=3$ in  Fig.~\ref{UNets}a).
\end{itemize}

Although some variants include spatial attention blocks \cite{G2}, we observed no significant performance gains and encountered training instability and increased computational cost. Therefore, attention-based models were excluded from our baseline comparisons.

\subsection{ConvNextU-Net (CNU-Net)}

CNU-Net (adapted from \cite{ohana2024well} where it showed good performance on several physical systems) is another modern U-Net variant that integrates ConvNext blocks \cite{liu2022convnet, xie2017aggregated}, which employ the ‘divide and conquer’ approach to broaden the spatial receptive fields and semantic representations without increasing computational cost. These blocks stack one channel-wise convolution \cite{chollet2017xception} with a 7×7 filter and
two sequential 1×1 pointwise convolutions. The latter expand the channel dimensions by a factor of 4 and contract them back.

\subsection{SineNet}

SineNet \cite{Z4}, Fig. \ref{UNets}b, addresses temporal misalignment in U-Net architectures by stacking multiple U-Nets sequentially, each operating at a reduced effective time step. Notably, each U-Net in the stack has internal skip connections, but no skip connections exist between encoder-decoder pairs across U-Nets. Thereby, deep (low-resolution) features are discarded between waves, limiting semantic continuity. Another feature is using average pooling for downsampling.

\subsection{DW-Net (ours)}

\subsubsection{Architectural Motivation}

In most U-Net variants, the network depth—defined as the total number of convolutional layers—is constrained by two factors. (1) The number of resolution levels is typically fixed at five based on empirical success across physical systems. (2) The number of convolutional layers per level is usually fixed at two. This leaves channel count (i.e., network width) as the primary tunable parameter. However, it is well-established that both depth and width must be carefully balanced to optimize performance. As shown in \cite{Bengio+chapter2007, larochelle2007empirical}, and supported by circuit complexity theory, shallow networks may require exponentially more units to match the expressiveness of deeper architectures, which scale polynomially. This motivates architectural designs that allow explicit control over depth, especially for multi-scale physical modeling.

\subsubsection{Insights from Multi-Scale Physical Dynamics}

Features of various scales emerging in complex systems interact predominantly locally. Small-scale features that are spatially separated do not interact directly, but can interact indirectly through larger-scale structures that encompass them. This makes U-Nets well-suited for multi-scale systems: it efficiently creates multi-resolution embeddings and captures local interactions. However, it allows only a single pass of cross-scale interaction, limiting the ability to progressively refine representations through network propagation.

\subsubsection{Improving Computational Efficiency}

In U-Net architectures (including SineNet), the highest-resolution layers—those with the largest spatial dimensions and lowest channel count—are typically the most computationally expensive. At the same time, these layers often encode less semantic information. DW-Net reduces their usage at intermediate waves, substantially improving efficiency without major sacrifices to accuracy.

\subsubsection{Design and Key Components of DW-Net}
DW-Net, Fig. \ref{UNets}c, is constructed by stacking multiple U-Net modules (waves), each with its own encoder--decoder pair. The key innovation lies in introducing \textbf{cross-wave skip connections} at matching resolution levels, enabling features to persist and evolve across waves. This facilitates hierarchical learning, repeated multiscale interactions, and progressive refinement of representations.

To improve computational efficiency, DW-Net omits the highest-resolution layers in intermediate waves---these layers are costly yet contribute less semantic information. Average pooling is used for downsampling, and transposed convolutions for upsampling. Each convolutional block is independently parameterized, avoiding weight sharing and allowing flexible learning.

DW-Net shares structural similarities with LadderNet~\cite{zhuang2018laddernet}, originally proposed for medical image segmentation. Both architectures employ intra-wave and cross-wave skip connections. However, DW-Net introduces several key differences:
\begin{itemize}
    \item \textbf{Aggregation method:} LadderNet uses summation; DW-Net uses channel-wise concatenation, preserving feature diversity.
    \item \textbf{Full-resolution recovery}: is avoided in intermediate waves, reducing cost.
    \item \textbf{Skip connectivity:} DW-Net includes skip connections at all resolution levels, including the deepest layers, which LadderNet omits.
    \item \textbf{Skip placement:} Skip connections are placed before concatenation, enhancing gradient flow and feature reuse.
    \item \textbf{Weight sharing:} DW-Net does not share weights across convolutional blocks, allowing more expressive learning.
    \item \textbf{Pooling:} Similar to SineNet, average pooling is used instead of stride-2 convolutions.
\end{itemize}

\subsection{U-Net\textsubscript{deep} (Deeper U-Net)}

To isolate the effect of multi-scale interactions, we also implement DeeperU-Net—a single-wave U-Net variant with deeper convolutional blocks (four layers per each block) and internal skip connections, as is illustrated at the bottom in Fig. \ref{UNets}. This variant does not use recurrent weight sharing, distinguishing it from R2U-Net \cite{alom2018recurrent} commonly used in vision applications.

\section{Physical systems and datasets}\label{Syst}

We evaluate all models on a diverse set of multi-scale fluid and plasma systems. Here, we provide high-level description of the physical systems, details including data generation and model training strategies are given in Appendix \ref{Systems}.

\subsection{Buoyant Incompressible Gas Flow with Smoke}\label{S-smoke}

This system represents thermal convection of light species, e.g., smoke, in a closed rectangular domain. The flow is governed by the incompressible Navier-Stokes equations augmented with a transport equation for the smoke concentration (assuming pure advection). The flow is driven by the buoyancy force which is proportional to the smoke concentration. 2D and 3D systems were modeled. The datasets adopted from \cite{G2} and \cite{L023} respectively, were generated using the $\Phi\text{Flow}$ solver \cite{holl2020phiflow}. In the 2D case, a 128×128 grid was used and the Reynolds number was 100. In the 3D case, a 64×64x64 grid was used and the Reynolds number was 333. A trajectory example is shown in Figs. \ref{Smoke-trajectory-d}, \ref{Smoke-trajectory-vx}, and \ref{Smoke-trajectory-vy} in Appendix \ref{roll}.

\subsection{Shallow-Water Planetary Atmosphere Model}\label{S-SW}

The shallow water (SW) equations are derived by depth-integrating the incompressible Navier-Stokes equations \cite{vreugdenhil2013numerical}. One of their applications is for modeling planetary atmospheres, predicting evolution of the pressure field (scalar) and wind velocity field (vector). We adopted the dataset from \cite{G2} for a model planet which was generated using a modified SpeedyWeather.jl \cite{klower2022milankl} solver on a cartesian 192 × 96 grid. A trajectory example is shown in Figs. \ref{SW-trajectory-p}, \ref{SW-trajectory-vx}, and \ref{SW-trajectory-vy} in Appendix \ref{roll}.

\subsection{Kolmogorov Flow Turbulence}

The 2D Kolmogorov flow problem is a common benchmark for studying developed fluid turbulence in periodic domains. The sinusoidal flow of viscous liquid is induced by a unidirectional periodic force and the dynamics is governed by the incompressible Navier-Stokes equations. The dataset adopted from \cite{L023} was generated using a modified pseudo-spectral solver with Re = 1000 and forcing factor f$_0$ = 8. The output resolution and time step were 256x256 and 1/16 respectively. A trajectory example is shown in Fig. \ref{K-omega} in Appendix \ref{roll}.

\subsection{Hasegawa–Wakatani (HW) Plasma Turbulence}

Hasegawa-Wakatani (HW) equations describe turbulence of fully-magnetized plasma in nuclear fusion devices. The model assumes that there a gradient in the plasma density transverse to an external uniform magnetic field. The dynamics is formulated for non-dimensional perturbations of plasma (ion) density $n$ and the electric potential $\phi$. Periodic boundary conditions are used. We have solved these equations for $n$ and $\phi$  using the BOUT++ code \cite{B}, for $\alpha$ = 0.01 and $\kappa$ = 0.5. Spatial resolution was 128x128 and the time step was 1. Trajectory example: Figs. \ref{HW-n} and \ref{HW-phi} in Appendix \ref{roll}.

\section{Experiments}\label{Experiments}

This section details the experimental setup designed to ensure a fair comparison between models and presents the results of the model comparison.

\subsection{Basis for Fair Model Comparison}

 We standardize key architectural and training hyperparameters where possible, and evaluate performance across a range of configurations to assess the trade-off between accuracy and computational cost. This approach enables a systematic and unbiased comparison of model capabilities across different architectures.

\subsubsection{Hyperparameter Selection}

To ensure fair comparisons, we standardized hyperparameters across models where possible:
\begin{itemize}
\item \textbf{Resolution levels}: All models use five levels, consistent with prior work \cite{G2,Z4,ohana2024well}.
\item \textbf{Channel expansion}: A fixed ratio of 2 between resolution levels is used throughout.
\item \textbf{Activation}: GELU activation function \cite{hendrycks2016gaussian} is used throughout.
\item \textbf{Boundary conditions (padding)}: consistently with \cite{G2}, periodic padding was used for
periodic domains; zero padding was used for other boundaries.
\end{itemize}

\subsubsection{Accuracy vs. Cost Trade-off}

Rather than comparing best-performing variants alone, we reconstruct the Pareto frontier of accuracy vs. computational cost. Computational cost is defined as training / inference time on a single A100-80GB GPU. Width (channel count) is varied for all models, whereas depth is additionally varied in ablation studies for SineNet and DW-Net.

We vary model width starting from 4 channels at the highest-resolution level, doubling for each bigger model. The upper limit on the model width was dictated by the training budgets capped at \mbox{8 h} (smoke, shallow-water), \mbox{5 h} (Kolmogorov), and \mbox{2000 s} (HW) wall clock time, all on a single A100--80GB GPU. 

\subsubsection{Training Setup}

Following \cite{G2,Z4}, all models were trained to predict one future state from a fixed number of past states (concatenated channel-wise), then rolled out auto-regressively to generate trajectories (details on the number of time steps in trajectories for each system are provided in Appendix \ref{Systems}).

All models were trained using the ADAM optimizer with a custom learning schedule featuring exponential decay with sinusoidal annealing (see Appendix \Ref{LR}). Batch size and epoch count were fixed per problem across models. Learning rate scaling was fine-tuned per model.

Scaled L2 loss was used to train model on the smoke, SW, and the Kolmogorov flows, whereas the mean squared error (MSE) was used for HW plasma turbulence. Each model was trained with 3 initializations (6 initializations for HW and Kolmogorov's flow) using fixed seeds $\{2,12,22,\dots\}$. We report the best-performing variant over the initializations.

\subsubsection{Evaluation protocol}

Following \cite{L023, Z4}, for smoke, shallow-water, and the Kolmogorov flows, we use the \emph{scaled L2} loss computed per time step and averaged across test trajectories:

\begin{equation}
L_{\text{2,t}}(\hat{u}t, u_t) = \frac{1}{M} \sum{k=1}^{M} \frac{|\hat{u}_t^k - u_t^k|_2}{|u_t^k|_2},
\label{eq:L2}
\end{equation}

where  $\hat{u}_t$ is the predicted field at time step $t$, 
$u_t$ is the ground truth field at time step $t$,
$M$ is the number of scalar fields,
$\hat{u}_t^k$ and $u_t^k$ are the $k$-th scalar fields of the prediction and ground truth, respectively,
$\|\cdot\|_2$ denotes the L2 norm over spatial dimensions. Trajectory examples are presented in Appendix \ref{roll}.

For HW turbulence, where the Lyapunov time ($\approx 0.5$ \cite{pedersen1996lyapunov}) is shorter than the output interval ($\Delta t = 1$), trajectory errors accumulate quickly (an example of a trajectory is shown in Figs. \ref{HW-n} and \ref{HW-phi} in Appendix); thus, we evaluate statistical fidelity over a single 2000-step rollout using time-averaged spatial FFT spectra an example of which is given in Fig. \ref{HW-graphs} in Appendix. Here we present aggregated errors for these quantities normalized by ground-truth variance:
\begin{equation}
\text{err}(y)=\frac{\langle (y-y_{\text{true}})^2\rangle}{\operatorname{var}(y_{\text{true}})},
\label{eq:err}
\end{equation}

where $y$ denotes the parameter of interest (FFT harmonics or autocorrelation), and  $\langle \cdot \rangle$ denotes the average over the x-axis in Fig. \ref{HW-graphs} ($k$ for spectra, time for autocorrelations).

\subsection{Pareto Trade-off Between Prediction Accuracy and Computational Cost}

In Sections \ref{Flows} and \ref{Plasma}, we benchmark the single-wave U-Net variants against \mbox{SineNet with 2 waves} and \mbox{DW-Net-3 (with 3 waves)}. SineNet was used with fewer waves due to higher computational cost. Then, in Section \ref{Waves}, we present the results of the ablation studies to quantify the effect of stacking additional waves in the DW-Net architecture. 

\subsubsection{Comparing on Buoyant Smoke, Shallow-Water and Kolmogorov's Flows}\label{Flows}

\begin{figure}[hbt!]
  \centering
  \scalebox{1}[0.9]{\includegraphics[width=1.0\textwidth]{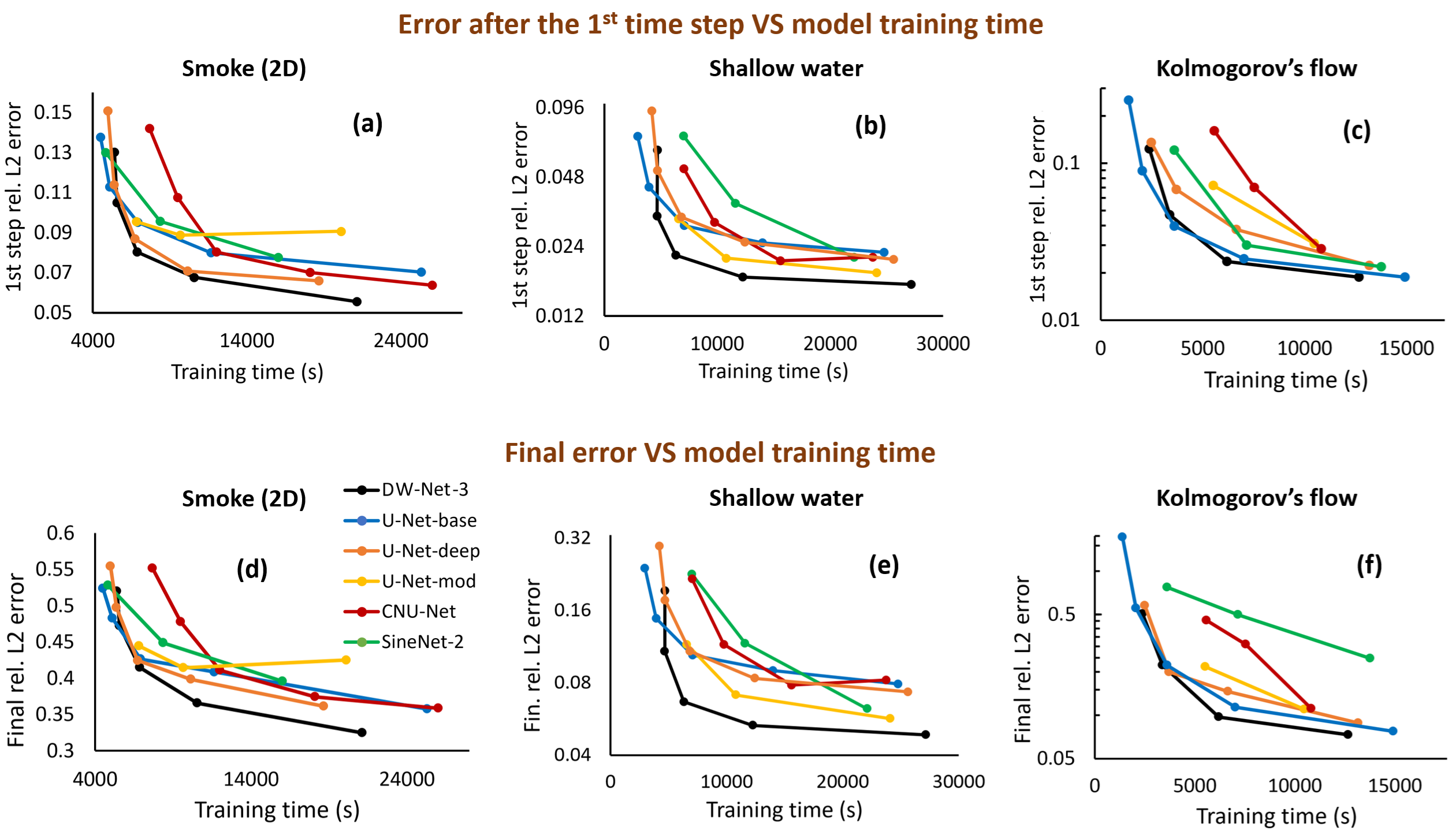}}
  \caption{Pareto frontiers illustrating the trade-off between prediction accuracy and training time (a proxy for computational cost) for various models applied to 2D buoyant smoke, shallow water, and Kolmogorov flow systems. Each curve corresponds to a model, obtained by varying its width, which results in different accuracy–cost operating points. Colors distinguish different model architectures. Curves closer to the bottom-left corner correspond to models with a more favorable trade-off. DW-Net-3 achieves the most favorable trade-off across the majority of benchmarks. Similar results for the trade-off between prediction accuracy and inference time are presented in Fig. \ref{Pareto2Dinf}.}
  \label{Pareto2D}
\end{figure}

Figure~\ref{Pareto2D} shows Pareto trade-offs between prediction error and training time for various models applied to 2D buoyant smoke, shallow water, and Kolmogorov flow systems. The models include several single-wave U-Net variants, a two-wave SineNet, and a three-wave DW-Net-3. Each curve corresponds to a single model, where the width is varied (e.g., 4, 8, 16 channels at the highest-resolution level), resulting in different accuracy-cost operating points.

Training time serves as a proxy for computational cost (all models were trained on a single A100-80GB GPU). Similar trends using inference time instead of training time are shown in Fig.~\ref{Pareto2Dinf} in the Appendix. Prediction errors are quantified as relative $L_2$ errors (\ref{eq:L2}) of the predicted 2D fields, evaluated both after the first time step and after the final time step (trajectory lengths vary by system; see Appendix \ref{Systems}).

Clear Pareto frontiers emerge for each model: increasing width reduces prediction error at the cost of higher computational expense. The error decreases rapidly as the number of channels increases from 4 to 16, followed by more gradual improvements for wider configurations.

DW-Net-3 achieves the most favorable trade-off across the majority of benchmarks. It consistently outperforms single-wave U-Net and SineNet baselines, reaching lower errors at lower cost and maintaining this advantage as cost increases. Compared to the best-performing baseline in each case, DW-Net-3 reduces error by approximately \(\sim\)30\% for the shallow water system, \(\sim\)20\% for the Kolmogorov flow, and \(\sim\)10\% for buoyant smoke. This corresponds to roughly 2–3× lower training time to achieve the same accuracy, with gains evident from the earliest rollout steps.

In contrast, baseline performance is case-dependent, with no single architecture consistently dominating. These results indicate that model depth is a key factor in improving the accuracy–cost trade-off, and that cross-wave skip connections play a critical role. Notably, U-Net\textsubscript{deep} performs best on 2D smoke and second-best on shallow water and Kolmogorov flow, suggesting that increased depth improves accuracy, but a single encoder–decoder pass remains insufficient.

\begin{figure}[hbt!]
  \centering
  \scalebox{1}[0.9]{\includegraphics[width=0.8\textwidth]{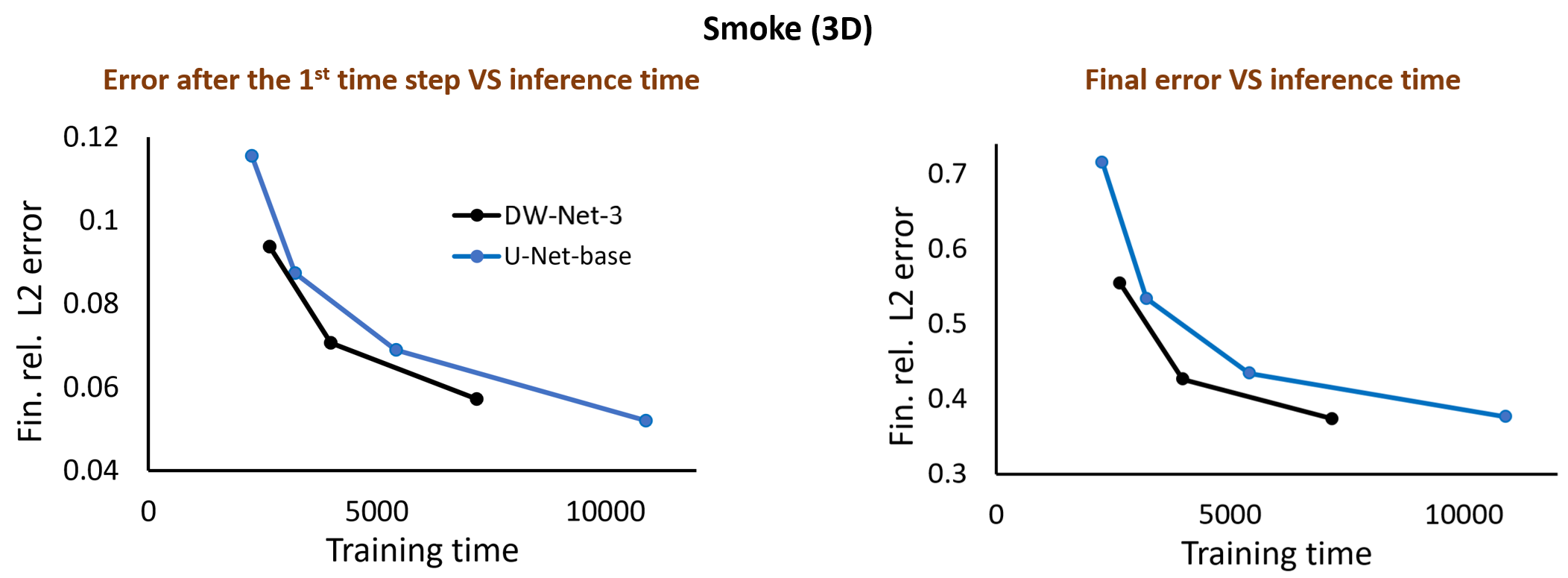}}
  \caption{Pareto frontiers illustrating the trade-offs between prediction accuracy and training/inference time for single-wave U-Net and 3-wave DW-Net models applied to the 3D buoyant smoke system. Performance improves toward the bottom-left corner, indicating lower error at lower training time. DW-Net achieves more favorable trade-off.}
  \label{Pareto3D}
\end{figure}

Limited 3D experiments were conducted on the buoyant smoke system, where the DW-Net-3 model is compared against \mbox{U-Net\textsubscript{base}} as the baseline. The results are presented in Fig.~\ref{Pareto3D}. Consistent with the 2D experiments, DW-Net-3 achieves a more favorable trade-off between accuracy and computational cost compared to the baseline model.

\subsubsection{Comparing on Hasegawa-Wakatani Plasma Turbulence (statistical fidelity)}\label{Plasma}

\begin{figure}[hbt!]
  \centering
  \scalebox{1}[0.9]{\includegraphics[width=0.9\textwidth]{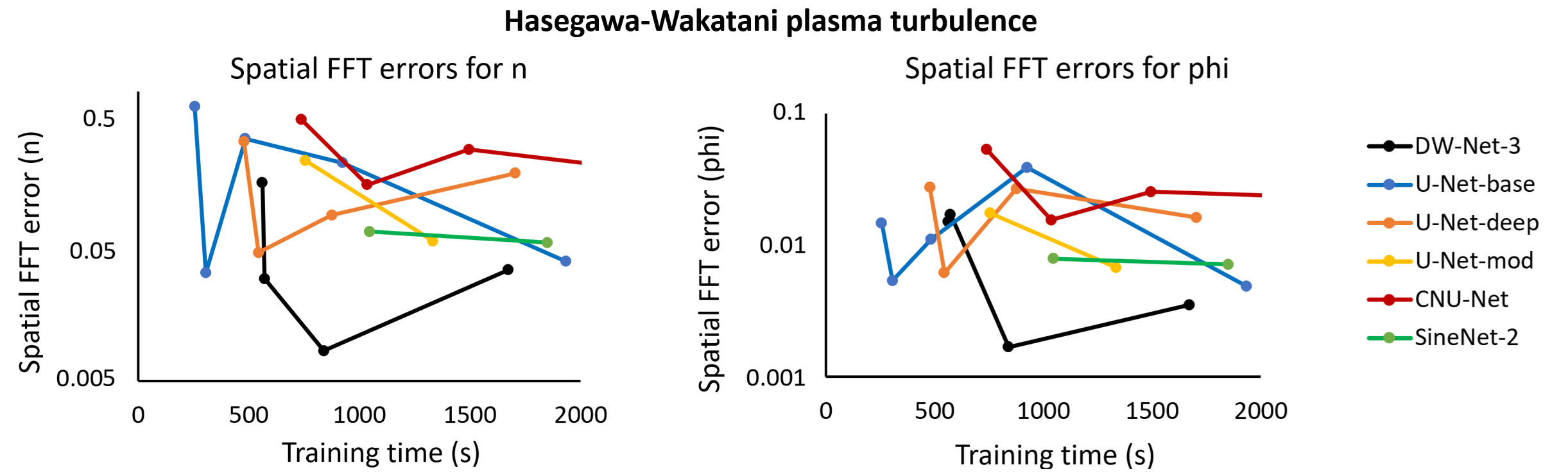}}
  \caption{Pareto trade-off between prediction accuracy (measured via statistical characteristics of time-averaged spatial FFT errors) for $n$ (left) and $\phi$ (right), and training time for various models applied to HW plasma turbulence. Each curve corresponds to a model, obtained by varying its width, yielding different accuracy--cost operating points. DW-Net-3 achieves the most favorable trade-off.}
  \label{ParetoHW}
\end{figure}

Figure~\ref{ParetoHW} shows Pareto trade-offs statistical prediction errors and training time for various models applied to 2D Hasegawa-Wakatani plasma turbulence. The models include several single-wave U-Net variants, a two-wave SineNet, and a three-wave DW-Net-3. Each curve corresponds to a single model, where the width is varied. Prediction errors are quantified using the overall error metric (\ref{eq:err}) for temporally-averaged FFT spectra of the density  \(n\) and electric potenteial $\phi$ fields, an example of which is given in Fig. \ref{HW-graphs} in Appendix.

The DW-Net-3 model outperforms other models by a large margin, with about an order of magnitude improvement for the FFT of \(n\) and more than an order-of-magnitude improvement for \(\phi\). Increasing width does not monotonically improve the performance for the HW setup; most models degrade at large widths.

\subsection{Ablation Study on the Number of Waves}\label{Waves}

\begin{figure}[hbt!]
  \centering
  \scalebox{1}[0.9]{\includegraphics[width=1.0\textwidth]{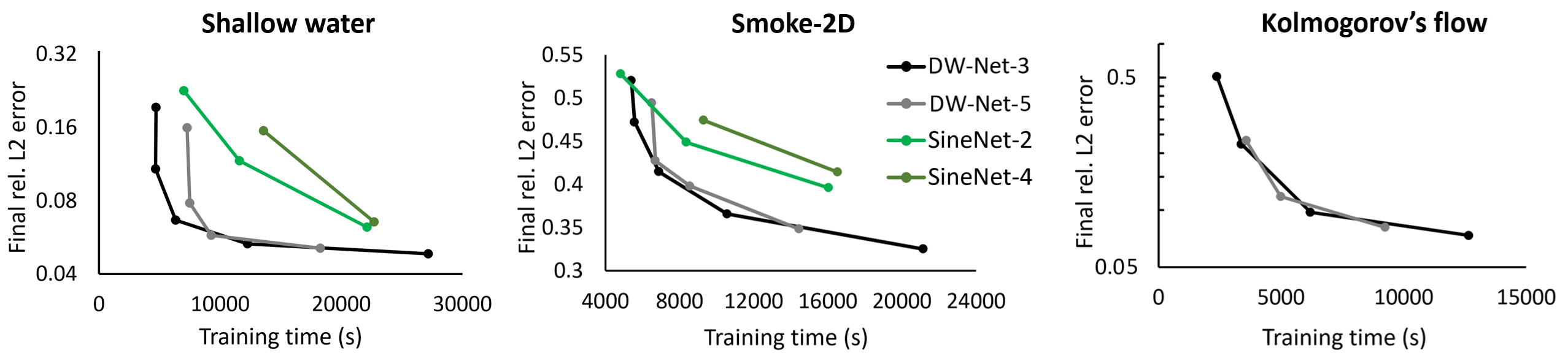}}
  \caption{Trade-offs between prediction error at the final trajectory step and training time for DW-Net and SineNet models with varying numbers of waves on 2D test systems. Additional results showing trade-offs for prediction error after the first time step against inference time are provided in Fig.~\ref{Ablation-add}.}
  \label{Ablation}
\end{figure}

We conducted ablation studies on 2D buoyant smoke, shallow-water, and Kolmogorov flows to assess the effect of increased model depth achieved by stacking additional U-Net waves. Models with two additional waves (DW-Net-5 and SineNet-4) are compared to those evaluated earlier (DW-Net-3 and SineNet-2). The results in Fig.~\ref{Ablation} show no noticeable improvement from extending the models beyond two waves for either DW-Net or SineNet. This suggests that, for these systems, two waves are sufficient to capture the relevant multi-scale dynamics, and that careful tuning of both model width and depth is required to achieve optimal performance.

\section{Conclusions}\label{Conclusion}

This work investigates the role of architectural capacity—specifically width and depth—in determining the performance of deep learning models for multi-scale physical dynamics. Rather than comparing models at fixed configurations, we adopt a Pareto-based evaluation framework that explicitly captures the trade-off between prediction accuracy and computational cost. Within this framework, we identify limitations of standard U-Net architectures, where depth is typically fixed, and introduce the Multi-Wave Network (DW-Net), which increases effective depth by stacking multiple encoder–decoder waves with both intra- and inter-wave skip connections.

Across a range of 2D and 3D benchmark systems, including buoyant smoke, shallow-water, Kolmogorov flow, and plasma turbulence, DW-Net consistently achieves more favorable accuracy–cost trade-offs than single-wave U-Net and related baselines. In particular, DW-Net reaches lower errors at reduced computational cost and can achieve comparable accuracy with up to 3× less training time under identical training conditions. Ablation studies further show that increasing depth via additional waves improves performance up to a point, after which diminishing returns are observed, highlighting the importance of jointly tuning model width and depth.

These results demonstrate that model depth is a critical and underexplored factor in optimizing neural surrogates for multi-scale physical systems. The proposed multi-wave design enables repeated cross-scale interactions and progressive refinement of representations, providing a flexible mechanism to control depth and improve efficiency without sacrificing accuracy.

\section{Future Work}

This work primarily focuses on autoregressive prediction tasks using high-fidelity simulated data. Extending DW-Net to settings involving noisy or experimental data, as well as data assimilation, remains an important direction for future research. While initial 3D experiments indicate promising scalability, broader evaluation across higher-dimensional and more complex systems is needed to fully assess performance. Additionally, DW-Net has so far been applied only to regular Cartesian grids; adapting the architecture to unstructured meshes via graph-based convolutions \cite{P1, K5, G22, G23, F2}, as demonstrated in U-Net-based graph neural networks \cite{J4, D24}, could significantly broaden its applicability to realistic geometries.

Finally, although DW-Net shares structural similarities with architectures used in computer vision, its potential beyond physical modeling, such as in semantic segmentation or other dense prediction tasks, remains unexplored. More broadly, the Pareto-based evaluation framework adopted in this work may provide a useful paradigm for fair and informative comparison of neural architectures in scientific machine learning, where both accuracy and computational efficiency are critical.

\section{Acknowledgement}

This material is based upon work supported by the U.S. Department of Energy, Office of Science, under Award Number DE-SC0024522.

\newpage
\appendix

\section{Details of the systems,  datasets, and model training}\label{Systems}
\subsection{The Kolmogorov flow}

The two-dimensional Kolmogorov flow problem is a benchmark for studying developed fluid turbulence in periodic domains. The sinusoidal flow of viscous liquid is induced by a unidirectional periodic force. The dynamics is governed by the incompressible Navier-Stokes equations in the vorticity form (non-dimensional):

\begin{align*}
\frac{\partial \omega(x,t)}{\partial t}
+ u(x,t)\cdot\nabla \omega(x,t)
&= \frac{1}{\mathrm{Re}}\,\nabla^{2}\omega(x,t) + f(x),
&& x \in (0,2\pi)^{2},\; t\in (0,T],\\[6pt]
\nabla\cdot u(x,t) &= 0,
&& x \in (0,2\pi)^{2},\; t\in [0,T],\\[6pt]
\omega(x,0) &= \omega_{0}(x),
&& x \in (0,2\pi)^{2}.\\[6pt]
\end{align*}

Here, $x$ and $y$ are spatial coordinates, $u$ is velocity (directed along $y$), $\omega$ is vorticity, $Re$ is the Reynolds number and $f$ represents the driving force along the y-direction \cite{kochkov2021machine}:

\begin{align*}
f(x) &= -f_0\cos(f_0 x) - 0.1\,\omega.
\end{align*}

The dataset adopted from \cite{L023} was generated using a modified pseudo-spectral solver \cite{kolmogorov_flow}, for $Re = 1000$ and with forcing factor $f_0$ = 8. The dataset comprises 100 trajectories for training and 20 trajectories for testing, where each trajectory has 160 states. Initial condition $\omega_0$ was sampled from a Gaussian random field following \cite{li2024physics}, ensuring a broad range of spatial scales in the flow.

The models were trained for 32 epochs to predict 1 time step ahead taking a single time step as input. The batch size was 20. Six initializations with fixed seeds were used for each model. The models were then rolled-out auto-regressively to generate trajectories of 16 time steps on a batch of 10 trajectories with randomly selected starting time frames. Relative L2 loss was used for both training and presenting the results.

The following learning rate scaling factors were used for the models:
U-Net\textsubscript{base} – 0.25, U-Net\textsubscript{mod} – 0.125, CNU-Net – 0.25, U-Net\textsubscript{deep} – 1, DW-Net – 0.25, SineNet – 0.25.

\subsection{Buoyant Incompressible Gas Flow with Smoke (2D and 3D)}

This system represents thermal convection of light species, e.g., smoke, in a closed domain. The flow is governed by the incompressible Navier-Stokes equations which assume that the flow velocity is too low to affect fluid density (Mach number \(<\)\(<\) 1, which is true for thermal convection). The equations are augmented by a transport equation for smoke concentration (assuming pure advection) and are solved in non-dimensional form:

\begin{align*}
\frac{\partial u(x,t)}{\partial t}
+ u(x,t)\cdot\nabla u(x,t)
&= \frac{1}{\mathrm{Re}} \nabla^{2} u(x,t) - \nabla p(x,t) + f(x,t),
&& x \in (0,L)^{n},\; t\in (0,T],\\[6pt]
\frac{\partial d(x,t)}{\partial t}
+ u(x,t)\cdot\nabla d(x,t)
&= 0,
&& x \in (0,L)^{n},\; t\in (0,T],\\[6pt]
\nabla\cdot u(x,t) &= 0,
&& x \in (0,L)^{n},\; t\in [0,T],\\[6pt]
u(x,0) &= 0,\quad d(x,0) = d_{0}(x),
&& x \in (0,L)^{n}.\\[6pt]
\end{align*}

Here, $u$ is the velocity vector, $p$ is the pressure, $d$ is the concentration of light-weight species (e.g., smoke), $f$ is the vertically-directed buoyancy force, which is proportional to the smoke concentration $d$ with a factor 0.5. $n$ denotes the number of spatial dimensions. 

Dirichlet boundary conditions are applied to the velocity, and Neumann conditions to the smoke concentration.

\subsubsection{2D case}

We use the dataset from \cite{G2}, generated using the $\Phi\text{Flow}$ solver \cite{holl2020phiflow} on a 128×128 grid with an output time step of 1.5. The domain size is 32×32, and the Reynolds number is 100. The dataset contains 5,200 training trajectories and 1,300 test trajectories, each with 14 time steps from randomly sampled initial conditions.

Following \cite{G2}, the models are trained to predict one time step ahead using the previous four time steps (concatenated channel-wise) as input. Models are then rolled out autoregressively to predict time steps 5–14. Each model is trained with 3 fixed-seed initializations (the best-performing realization is used). The error metric was the scaled L2 loss computed per time step, used for both training and evaluation. The models were trained for 80 epochs with batches of 40 time steps. A batch of 30 test trajectories was randomly selected for evaluation. The  errors for the first and the last time step are  presented, averaged across the test trajectories.

The following learning rate scaling factors were used for the models:
U-Net\textsubscript{base} – 1, U-Net\textsubscript{mod} – 0.25, CNU-Net – 1, U-Net\textsubscript{deep} – 1, DW-Net – 1, SineNet – 0.125.

\subsubsection{3D case}

To assess model scalability, we include a 3D version of the smoke flow system. The dataset, adopted from \cite{L023}, was also generated using the  $\Phi\text{Flow}$ solver on a 64×64×64 grid with a time step of 0.75 and Reynolds number of 333. The relative L2 loss was also used for training and evaluation. 

The dataset consists of 2,000 training trajectories and 200 test trajectories, each with 20 time steps. Models were trained for 10 epochs with a batch size of 20. 3D convolutions use the same filter sizes and channel expansion ratio (i.e., ×2) as in 2D. Performance is compared against the U-Net\textsubscript{base} baseline.

A learning rate scaling factor of 1 was used for both models.

\subsection{Shallow-Water Planetary Atmosphere Model}

The shallow water (SW) equations are derived by depth-integrating the incompressible Navier-Stokes equations \cite{vreugdenhil2013numerical}. One of their applications is for modeling planetary atmospheres, predicting evolution of the pressure field (scalar) and wind velocity field (vector). We adopted the dataset from \cite{G2} for a model planet, generated using a modified SpeedyWeather.jl \cite{klower2022milankl} solver. A cartesian grid 192 × 96 was used in combination with a fixed output time step of 48 h. 

The training and test data consisted of 5,600 and 1,400 trajectories respectively, each trajectory having 11 time steps. The models were trained for 80 epochs to predict one time step ahead (for time steps 3-11) using two previous time steps as input. The batch size was 36. Three initializations with fixed seeds were used for each model. The model was then run autoregressively to predict 9 time steps (3 to 11).  A batch of 30 trajectories were randomly selected from the test set to produce the presented results.

The following learning rate scaling factors were used for the models:
U-Net\textsubscript{base} – 0.25, U-Net\textsubscript{mod} – 0.25, CNU-Net – 0.25, U-Net\textsubscript{deep} – 1, DW-Net – 1, SineNet – 0.25.

\subsection{Hasegawa-Wakatani Plasma Turbulence}

The Hasegawa-Wakatani (HW) equations \cite{H83} describe turbulence relevant to fully-magnetized plasma in nuclear fusion devices. The model assumes a gradient in plasma density transverse to an external uniform magnetic field. The equations are formulated for normalized (non-dimensional) perturbations of plasma (ion) density $n$ and electric potential $\phi$ ($n$ is normalized to the background plasma density, and $\phi$ is normalized to the electron temperature): 

\begin{align*}
\frac{\partial n}{\partial t} + \{\phi, n\} + \kappa \frac{\partial \phi}{\partial y}
&= \alpha (\phi - n) - D_n \nabla^4 n, \\
\frac{\partial}{\partial t} \Delta \phi + \{\phi, \Delta \phi\}
&= \alpha (\phi - n) - D_p \nabla^4 \phi.
\end{align*}

Here, $x$ and $y$ are the spatial coordinates (the background density gradient is in the $x$ direction). $\kappa$ and $\alpha$ are non-dimensional parameters representing the density gradient and plasma adiabaticity. $D_n$ and $D_p$ are hyper-diffusivity parameters added for numerical stability. The Poisson bracket in the HW equations is defined as: 

\begin{align*}
\{A, B\}=\frac {\partial A} {\partial x} \frac {\partial B}{\partial y}-\frac {\partial A} {\partial y} \frac {\partial B}{\partial x}.
\end{align*}

Periodic boundary conditions are used. 

We solve these equations for $n$ and $\phi$  using the BOUT++ code \cite{B}, for $\alpha$ = 0.01 and $\kappa$ = 0.5. The hyper-diffusivity parameters were set to small values, $D_n = D_p = 0.0001$, to ensure numerical stability without affecting the results. Computations were performed on a high-performance computing cluster utilizing eight A100 GPUs. Spatial resolution was 128x128 with a time step of 1. The solver completed the task in approximately 3 hours. A single trajectory was modeled, initiated from white noise. The first 500 time steps corresponded to the warm-up stage, followed by instability growth and saturation. The subsequent 4,300 time steps corresponded to developed (quasi-steady) turbulence. Of those time steps, 4,000 were used for training and 300 for testing.

The models were trained to predict 1 time step ahead using 1 time step as an input. Batch size was 40, with 160 training epochs. The models were then rolled-out auto-regressively to generate trajectories of 2000 time steps. Since the Lyapunov time for HW turbulence is about 0.5 \cite{pedersen1996lyapunov}, i.e., smaller than the output timestep, we compare statistical characteristics of the generated turbulence to the ground truth (no tracing of individual trajectories).

\section{Learning Schedule}\label{LR}

A custom learning rate $lr$ schedule was applied for all models, based on a warm-up stage followed by an exponential decay combined with cosine annealing \cite{L17}, as determined by the following expression:
\[
lr = 0.01 \cdot \alpha \cdot \exp\left(-5\frac{\max(i, N_{\text{warm}}) - N_{\text{warm}}}{N_{\text{total}}}\right)
\cdot \left(0.8 + 0.5 \cdot \sin\left(2\pi\left(0.75 + \frac{i}{N_{\text{warm}}}\right)\right)\right).
\]
Here,  $i$ is the epoch number, $N_{\text{total}}$ is the total number of epochs reserved for learning, $N_{\text{warm}}=N_{\text{total}}/2$ corresponds to the linear warm-up stage, $\alpha$ is the scaling factor, which was fine-tuned in the range 0.125 - 1.0 for each model.

\newpage

\section{Accuracy vs. Computational Cost Trade-off (Additional Results)}

\begin{figure}[hbt!]
  \centering
  \scalebox{1}[0.9]{\includegraphics[width=1.0\textwidth]{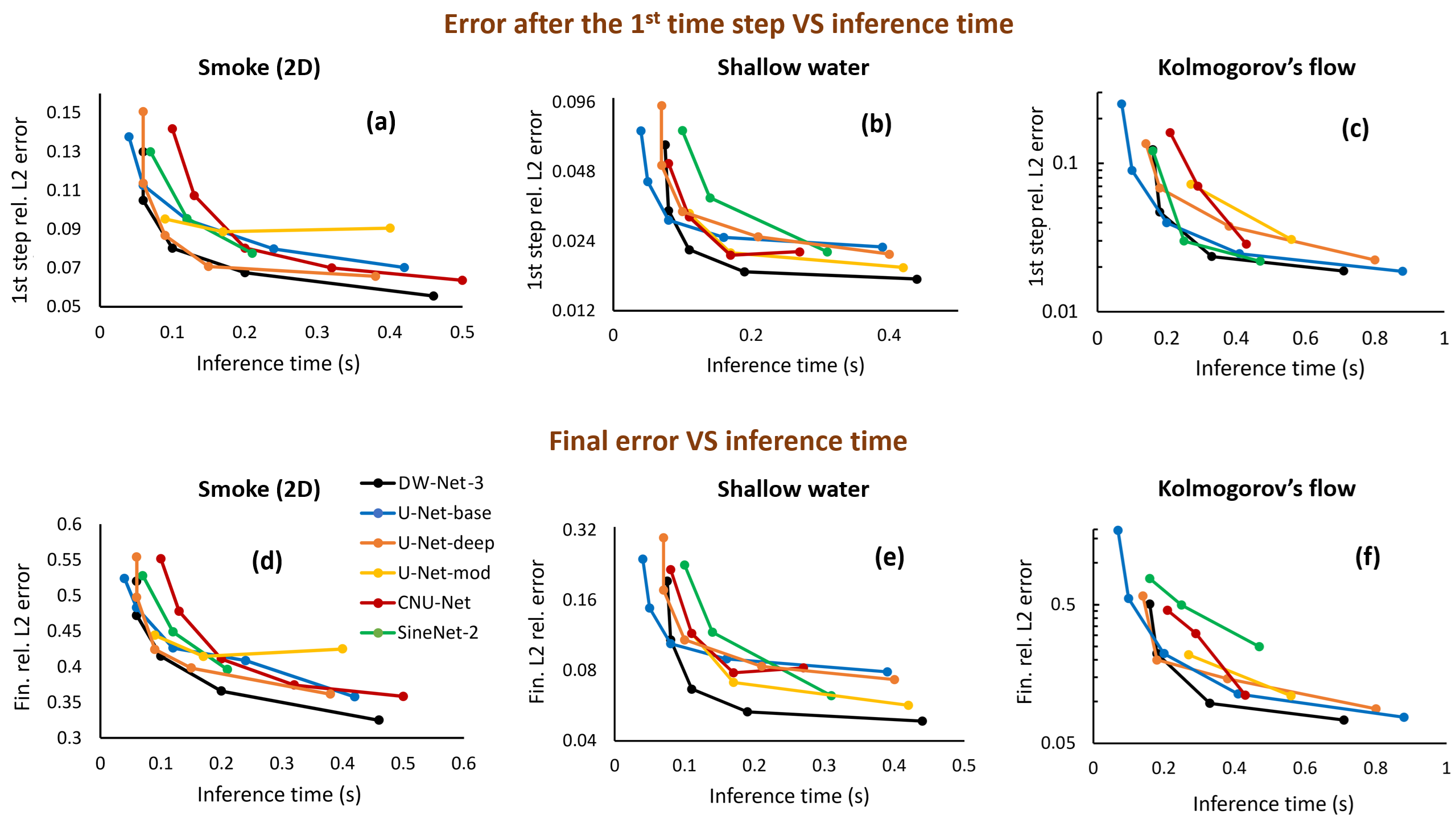}}
  \caption{Pareto frontiers illustrating the trade-off between prediction accuracy and inference time (a proxy for computational cost) for various models applied to 2D buoyant smoke, shallow water, and Kolmogorov flow systems. Each curve corresponds to a model, obtained by varying its width, which results in different accuracy–cost operating points. Colors distinguish different model architectures. Performance improves toward the bottom-left corner, indicating lower error at lower training time. DW-Net achieves the most favorable trade-off across the majority of benchmarks.}
  \label{Pareto2Dinf}
\end{figure}

\begin{figure}[hbt!]
  \centering
  \scalebox{1}[0.9]{\includegraphics[width=0.5\textwidth]{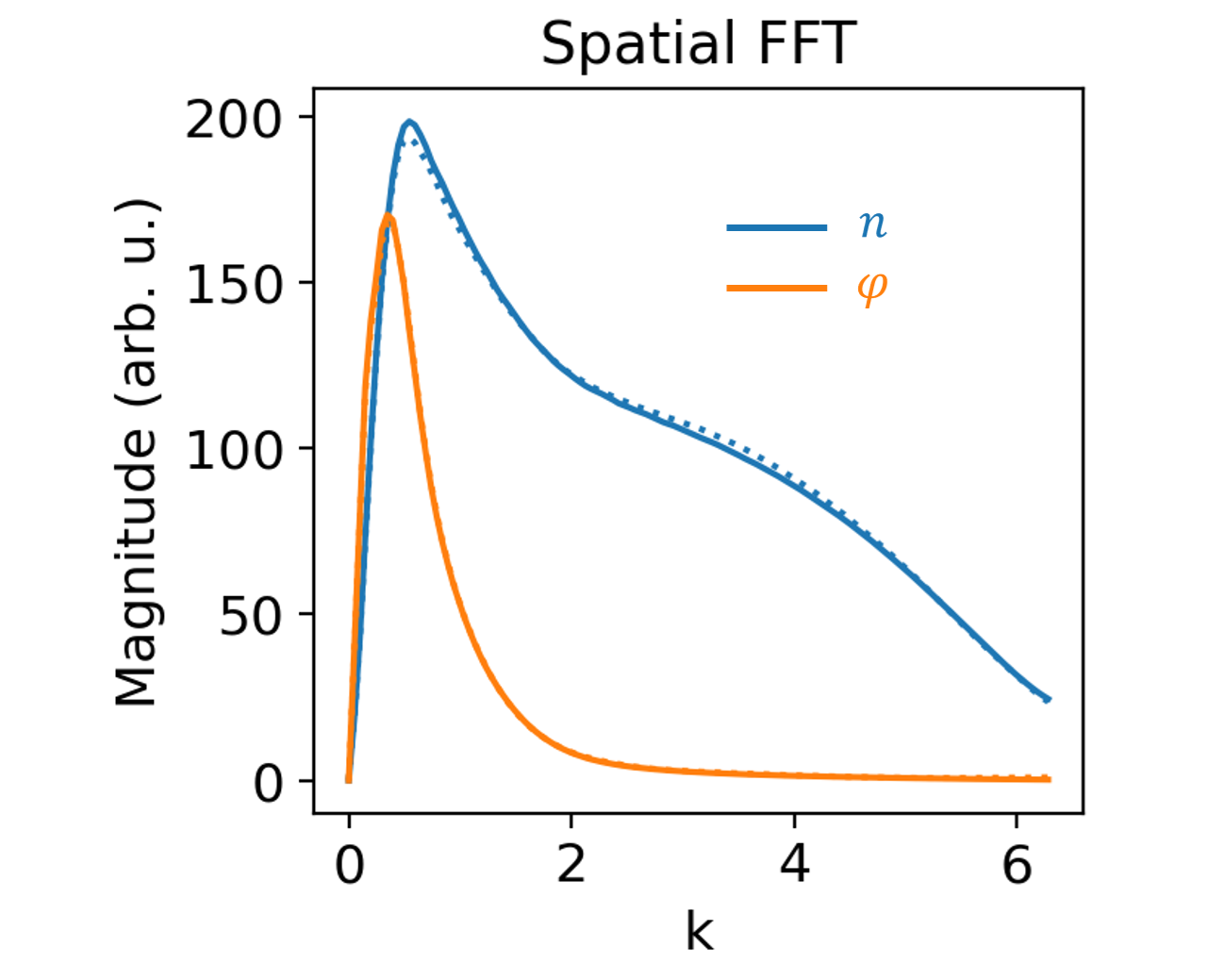}}
  \caption{Time-averaged spatial FFT spectra for $n$ and $\phi$. Solid lines -- ground truth (simulation data), dotted lines -- results of the DW-Net-3 model.}
  \label{HW-graphs}
\end{figure}

\begin{figure}[hbt!]
  \centering
  \scalebox{1}[0.9]{\includegraphics[width=1.0\textwidth]{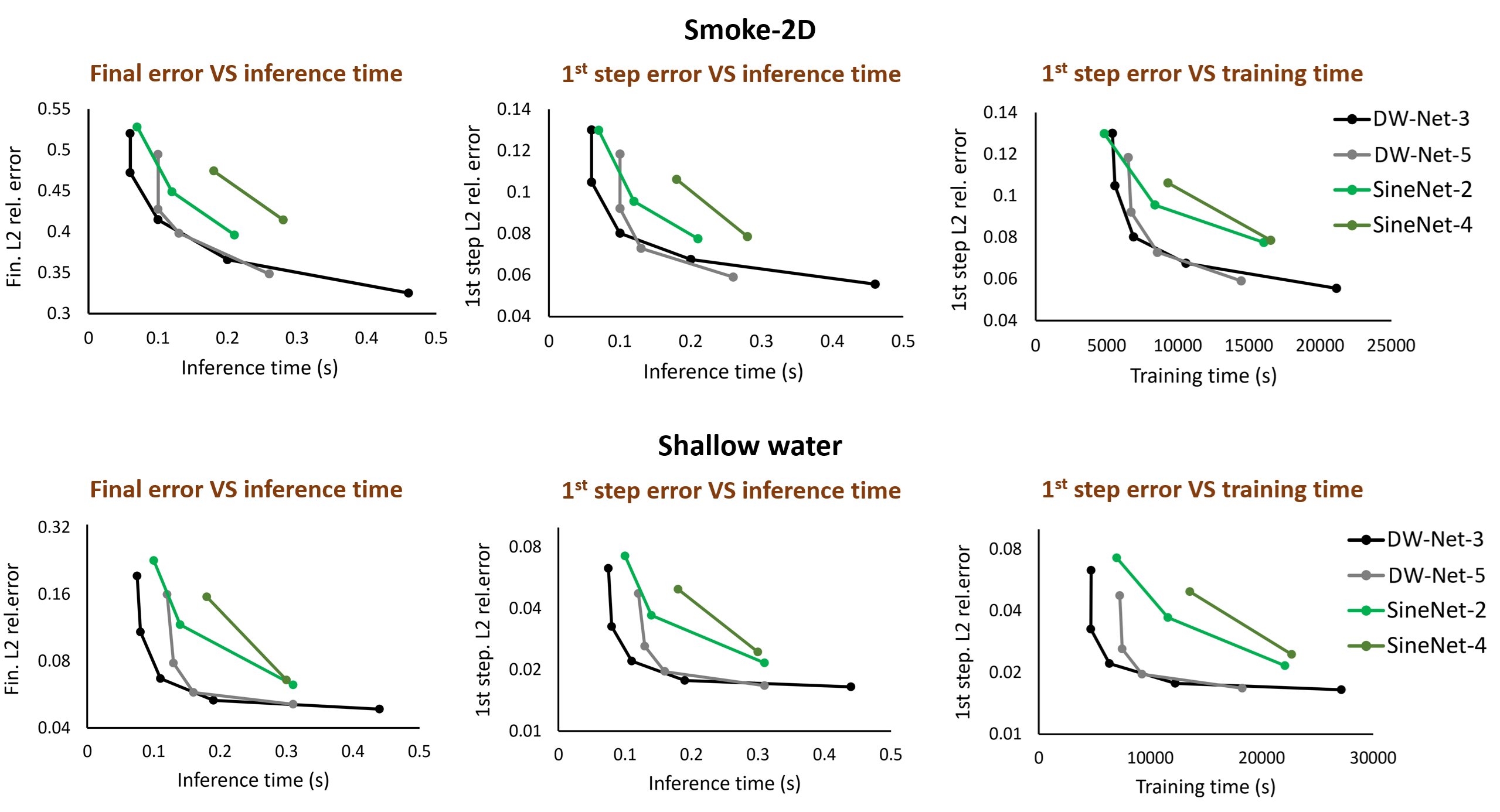}}
  \caption{Additional comparison results of DW-Net and SineNet models with more waves on two systems.}
  \label{Ablation-add}
\end{figure}

\clearpage

\section{Model rollouts}\label{roll}

\subsection{Buoyant flow of smoke}

\begin{figure}[hbt!]
  \centering
  \includegraphics[width=0.95\textwidth]{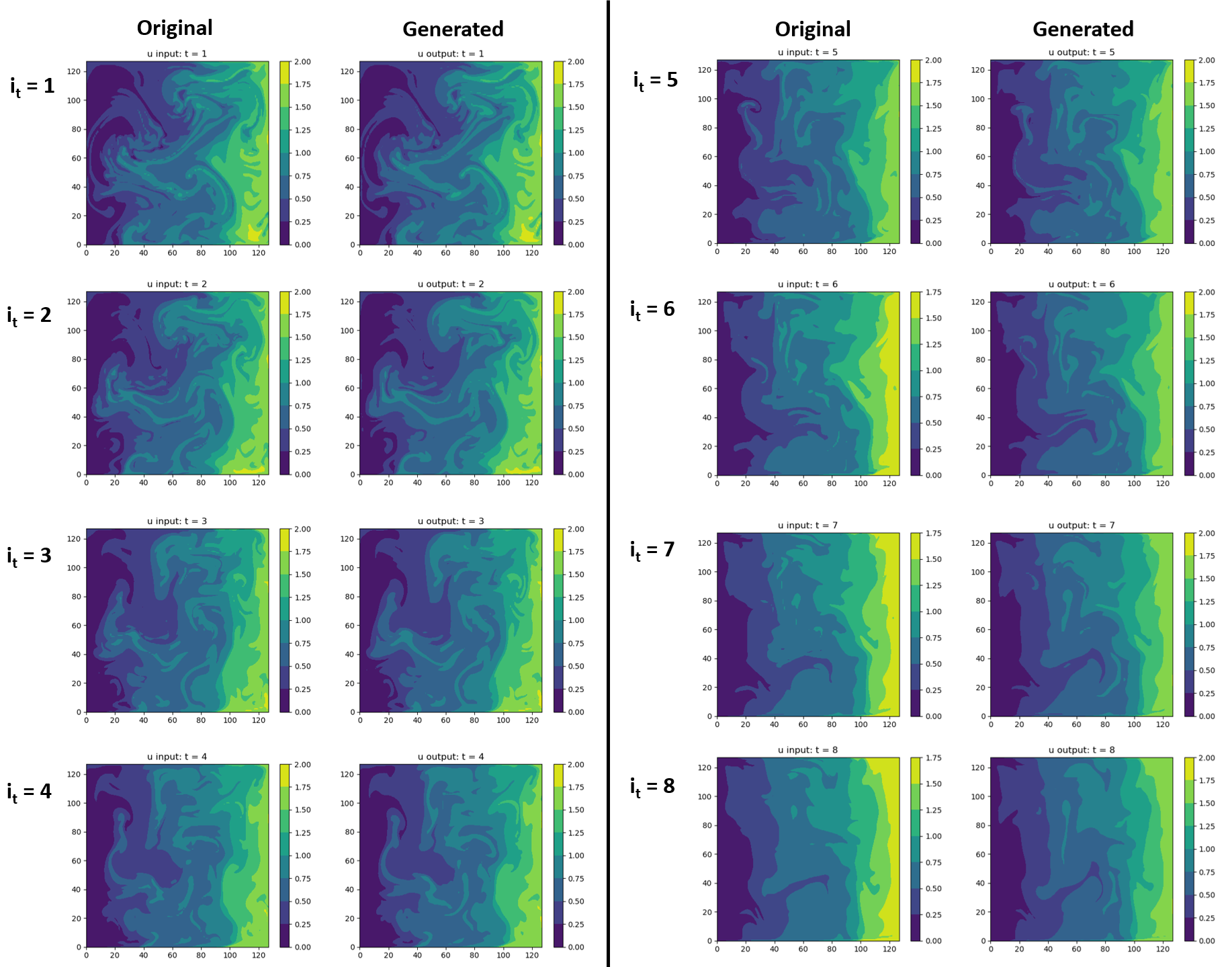}
  \caption{Example of a trajectory for the buoyant smoke flow generated by the DW-Net-3 model (best realization). The field of smoke density.}
  \label{Smoke-trajectory-d}
\end{figure}

\begin{figure}[hbt!]
  \centering
  \includegraphics[width=0.95\textwidth]{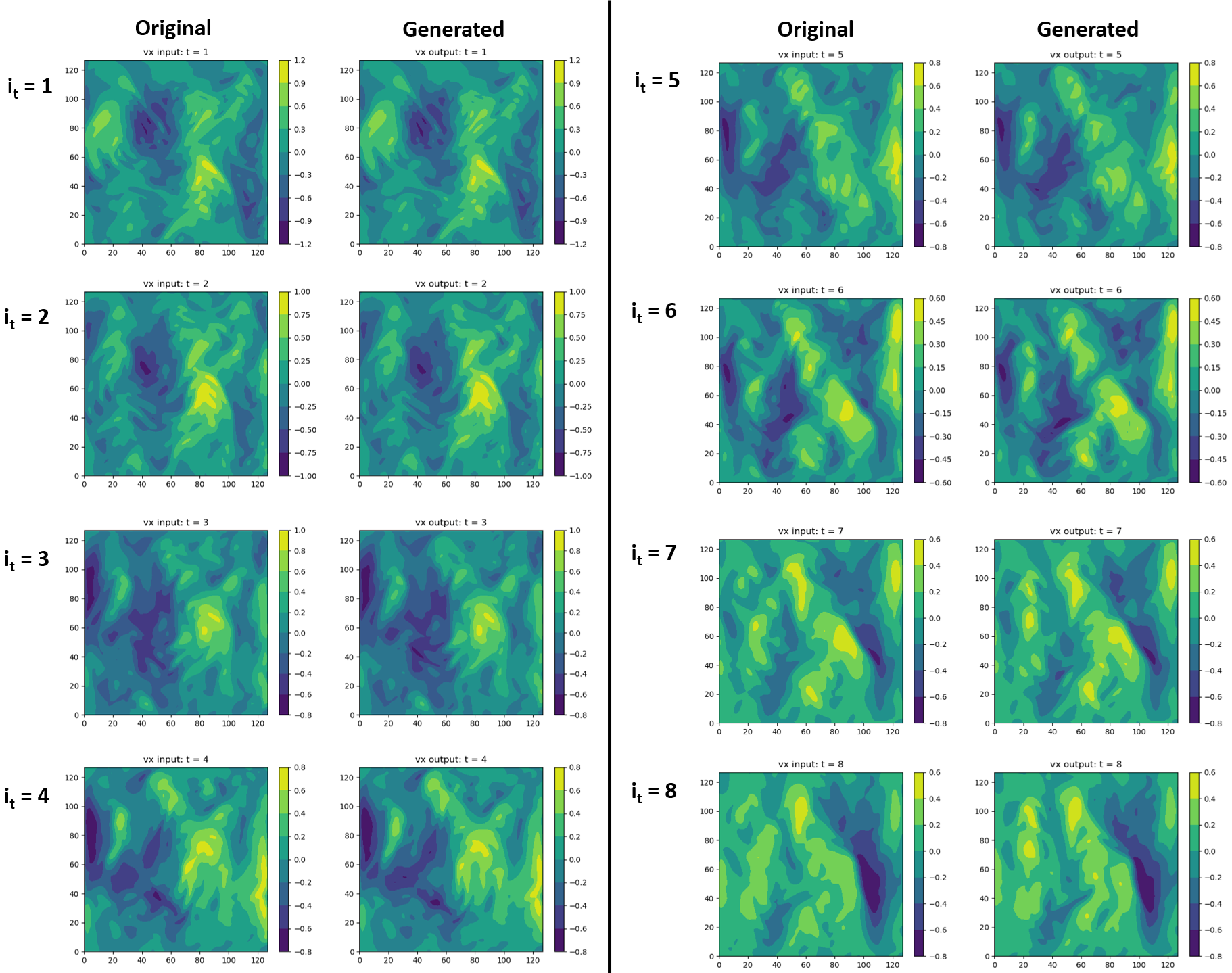}
  \caption{Example of a trajectory for the buoyant smoke flow generated by the DW-Net-3 model (best realization). The field of velocity (x-component).}
  \label{Smoke-trajectory-vx}
\end{figure}

\begin{figure}[hbt!]
  \centering
  \includegraphics[width=0.95\textwidth]{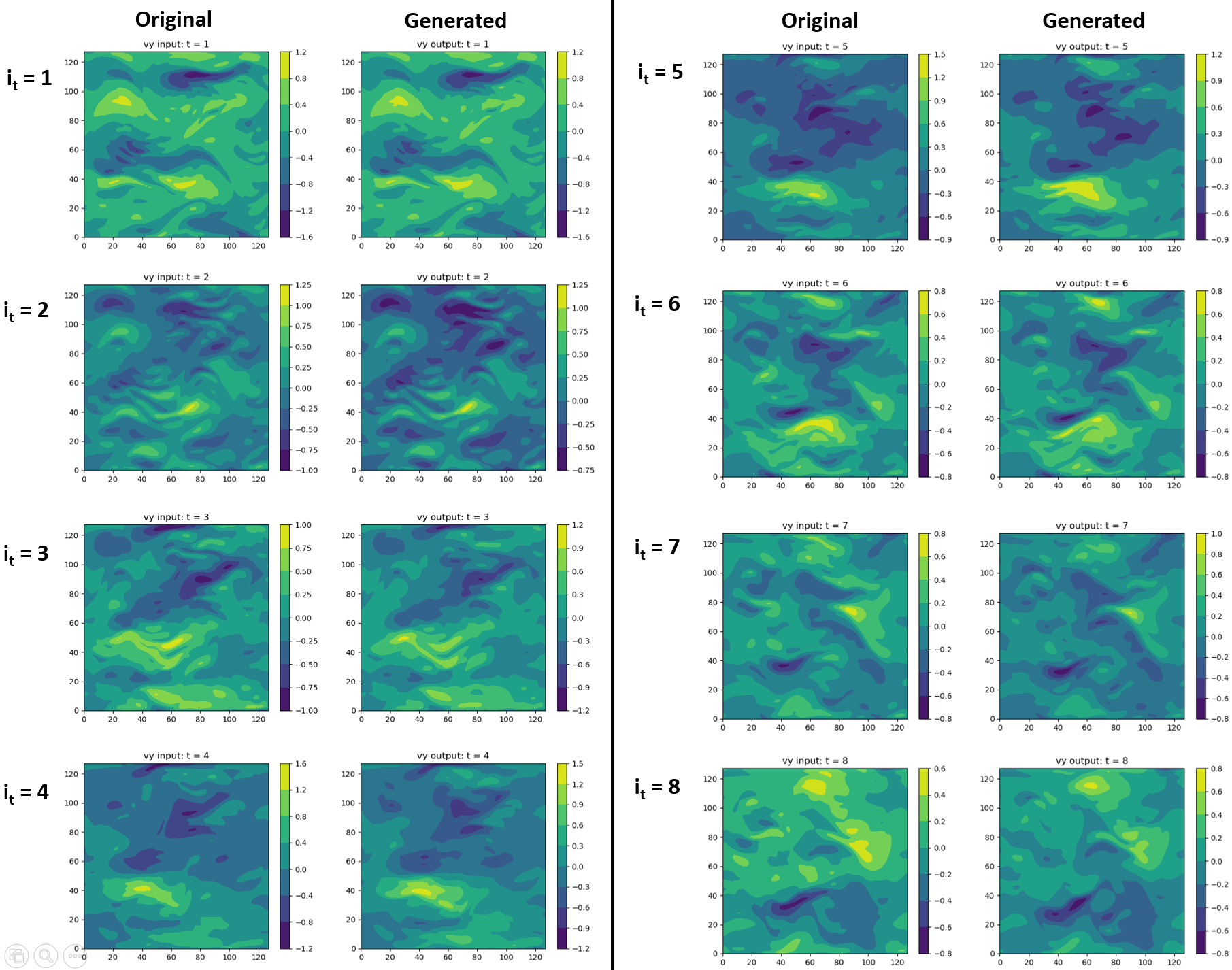}
  \caption{Example of a trajectory for the buoyant smoke flow generated by the DW-Net-3 model (best realization). The field of velocity (y-component).}
  \label{Smoke-trajectory-vy}
\end{figure}

\clearpage

\subsection{The Shallow Water System}

\begin{figure}[hbt!]
  \centering
  \includegraphics[width=0.95\textwidth]{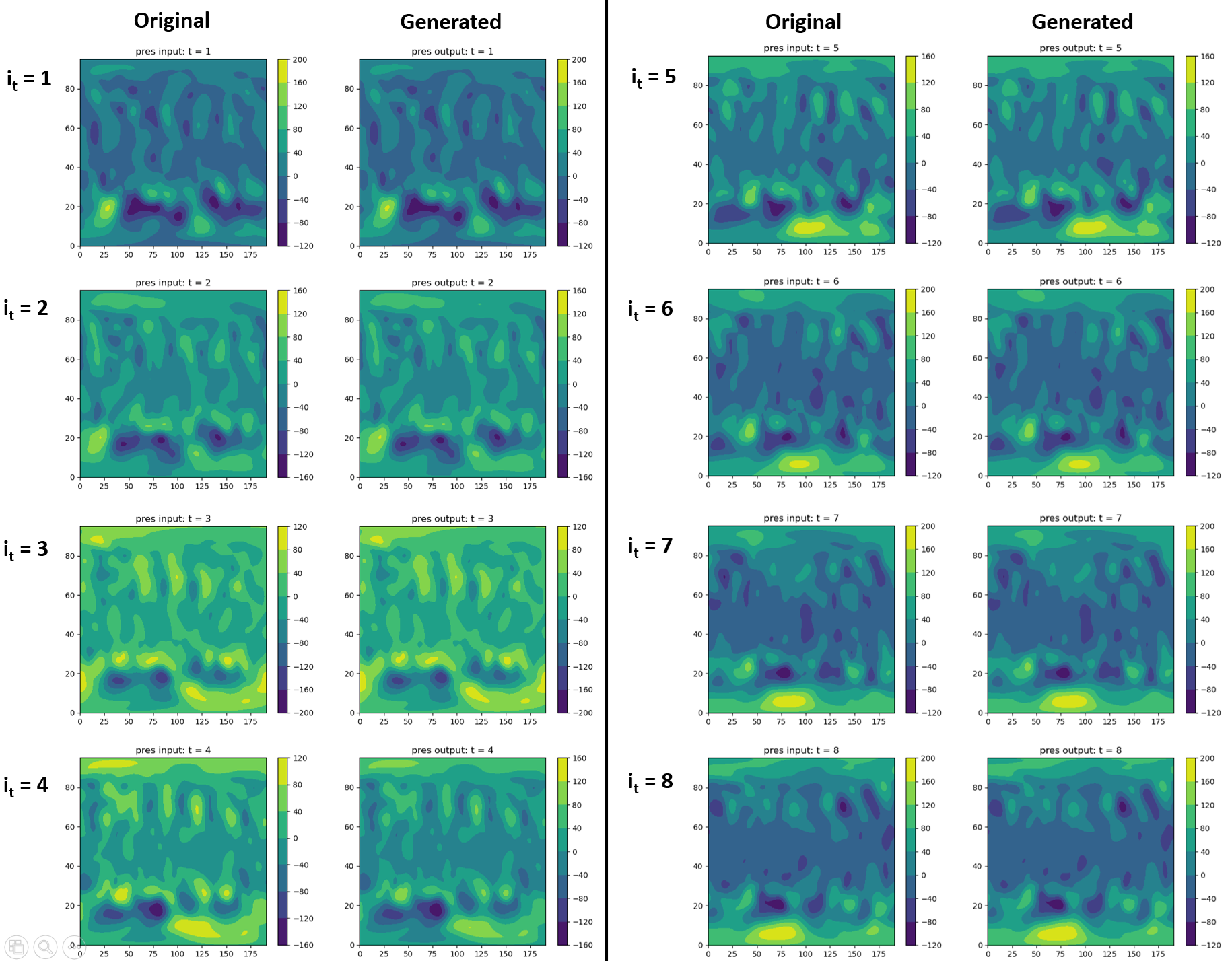}
  \caption{Example of a trajectory for the shallow water system generated by the DW-Net-3 model (best realization). The field of pressure.}
  \label{SW-trajectory-p}
\end{figure}

\begin{figure}[hbt!]
  \centering
  \includegraphics[width=0.95\textwidth]{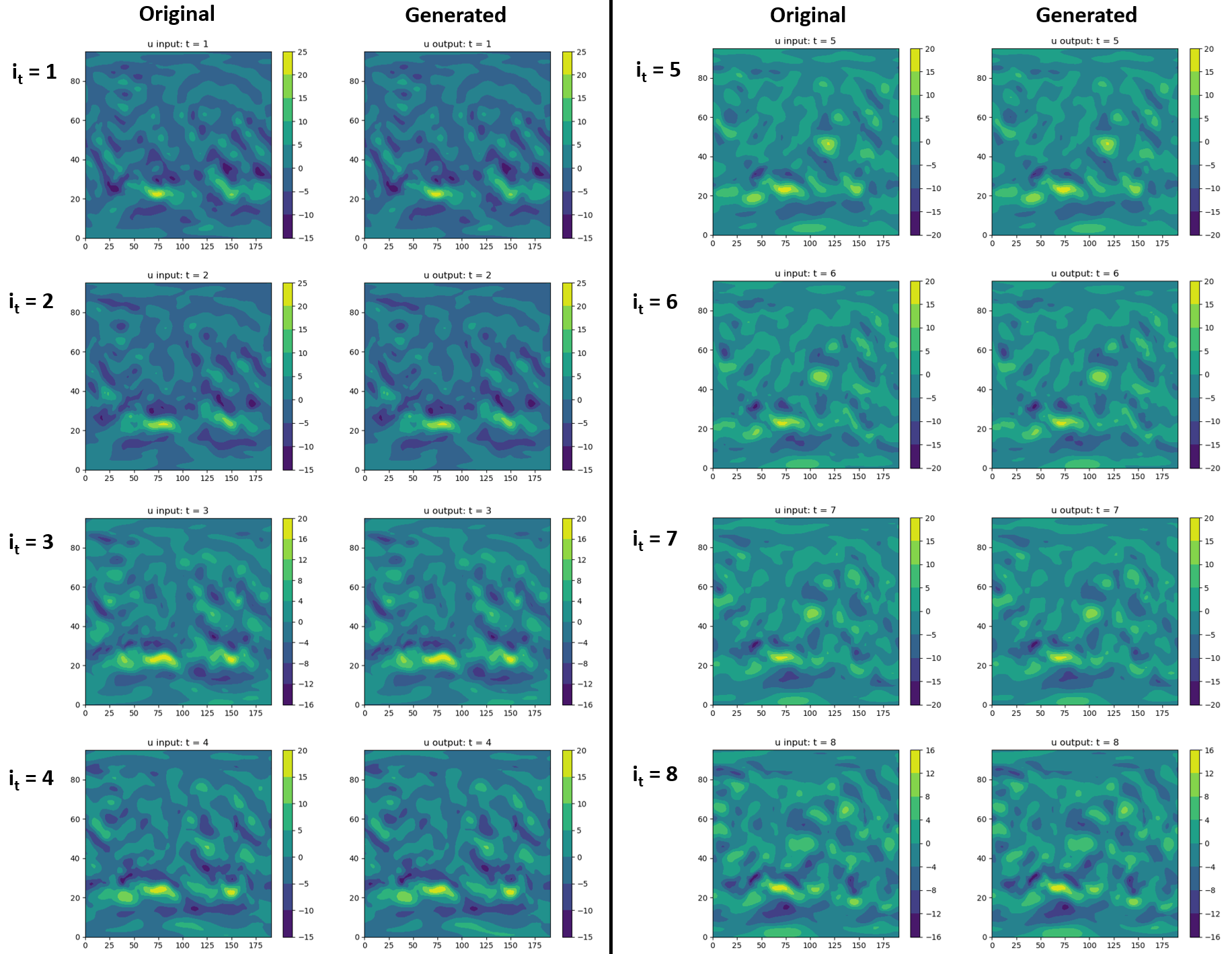}
  \caption{Example of a trajectory for the shallow water system generated by the DW-Net-3 model (best realization). The field of velocity (x-component).}
  \label{SW-trajectory-vx}
\end{figure}

\begin{figure}[hbt!]
  \centering
  \includegraphics[width=0.95\textwidth]{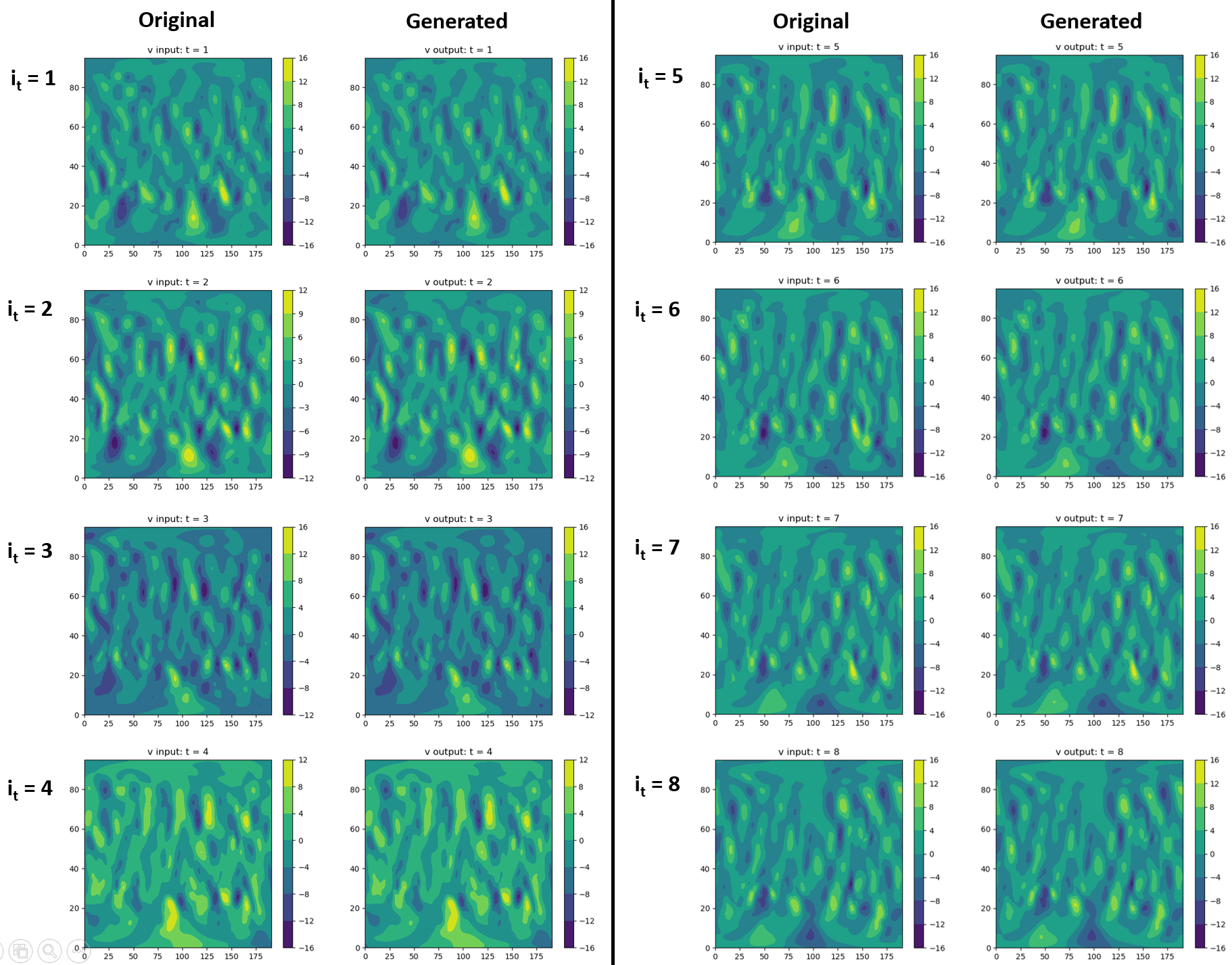}
  \caption{Example of a trajectory for the shallow water system generated by the DW-Net-3 model (best realization). The field of velocity (y-component).}
  \label{SW-trajectory-vy}
\end{figure}

\clearpage

\subsection{Kolmogorov's flow}

\begin{figure}[hbt!]
  \centering
  \includegraphics[width=0.95\textwidth]{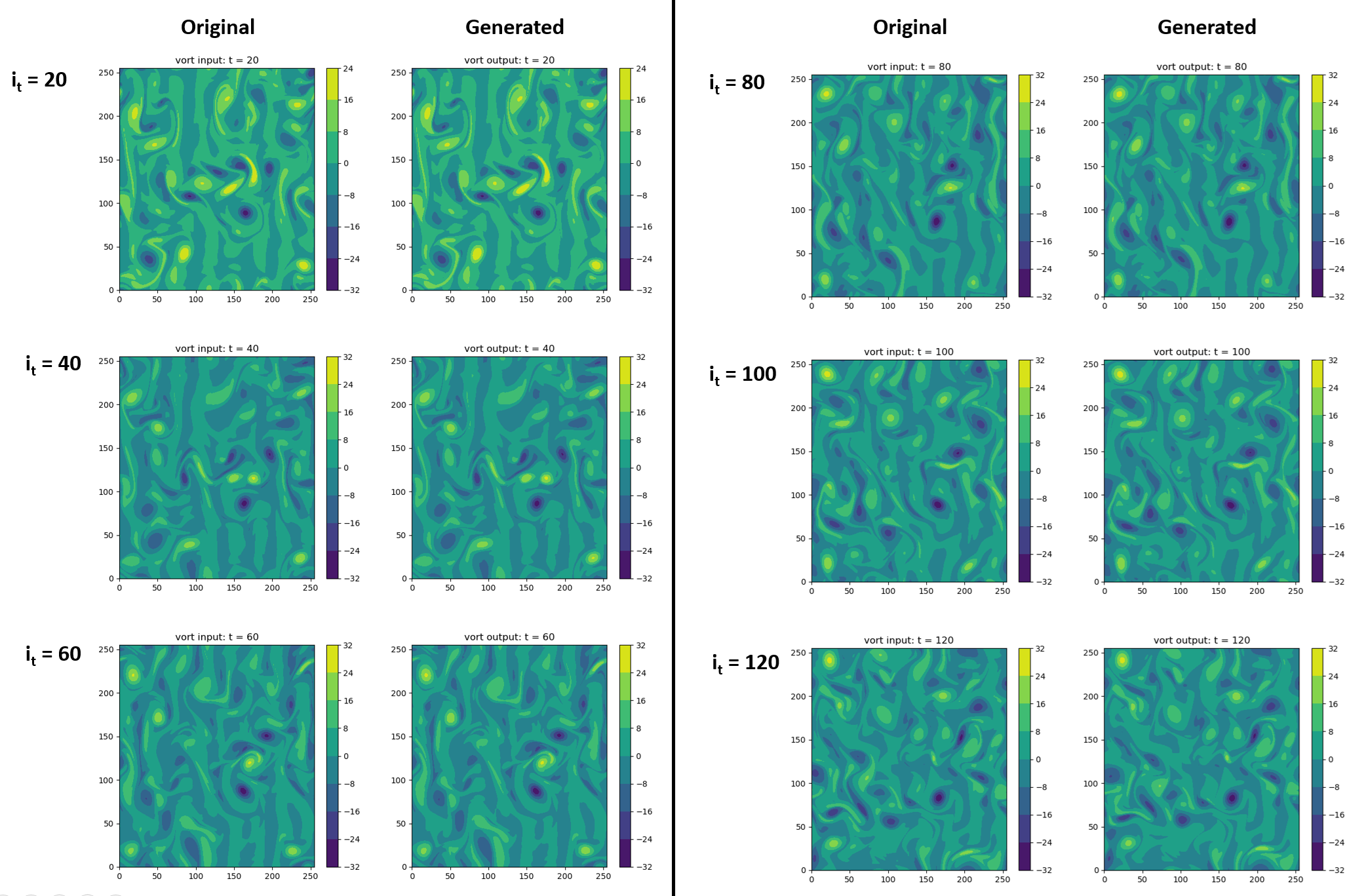}
  \caption{Example of a trajectory for the Kolmogorov turbulence generated by the DW-Net-3 model (best realization). The field of vorticity. Good agreement persists until the end of the trajectory of 120 time steps.}
  \label{K-omega}
\end{figure}

\clearpage

\subsection{Hasegawa-Wakatani turbulence}

\begin{figure}[hbt!]
  \centering
  \includegraphics[width=0.95\textwidth]{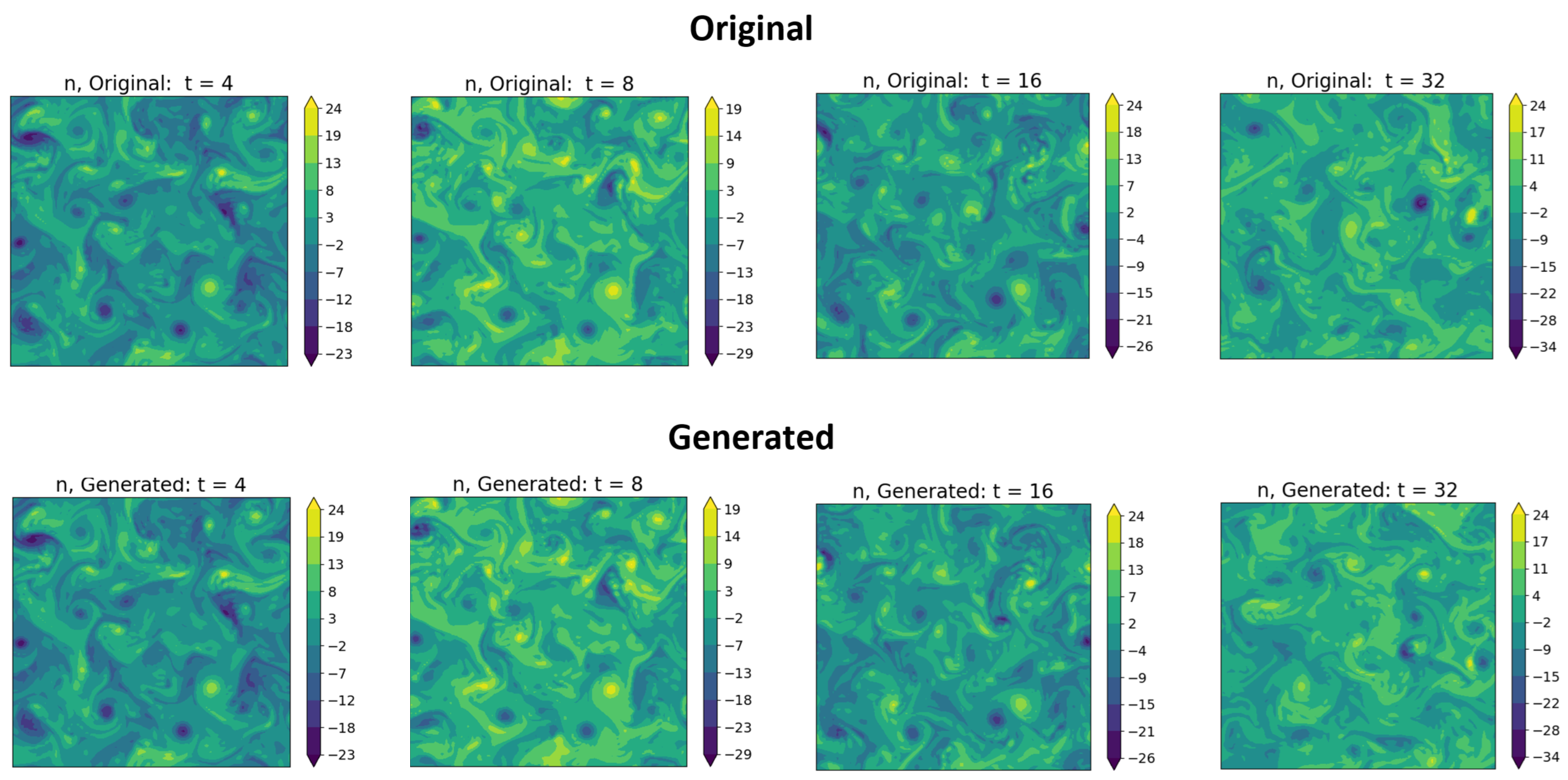}
  \caption{Beginning of a 2000 time step trajectory for HW turbulence generated by the DW-Net-3 model (best realization). The field of $n$: Generated data - bottom row vs. numerical simulation (using BOUT++) - top row. At first, the generated solution resembles the original (simulated) one quite closely. However, with time, the differences amplify and by the time step 32 become significant.}
  \label{HW-n}
\end{figure}

\begin{figure}[hbt!]
  \centering
  \includegraphics[width=0.95\textwidth]{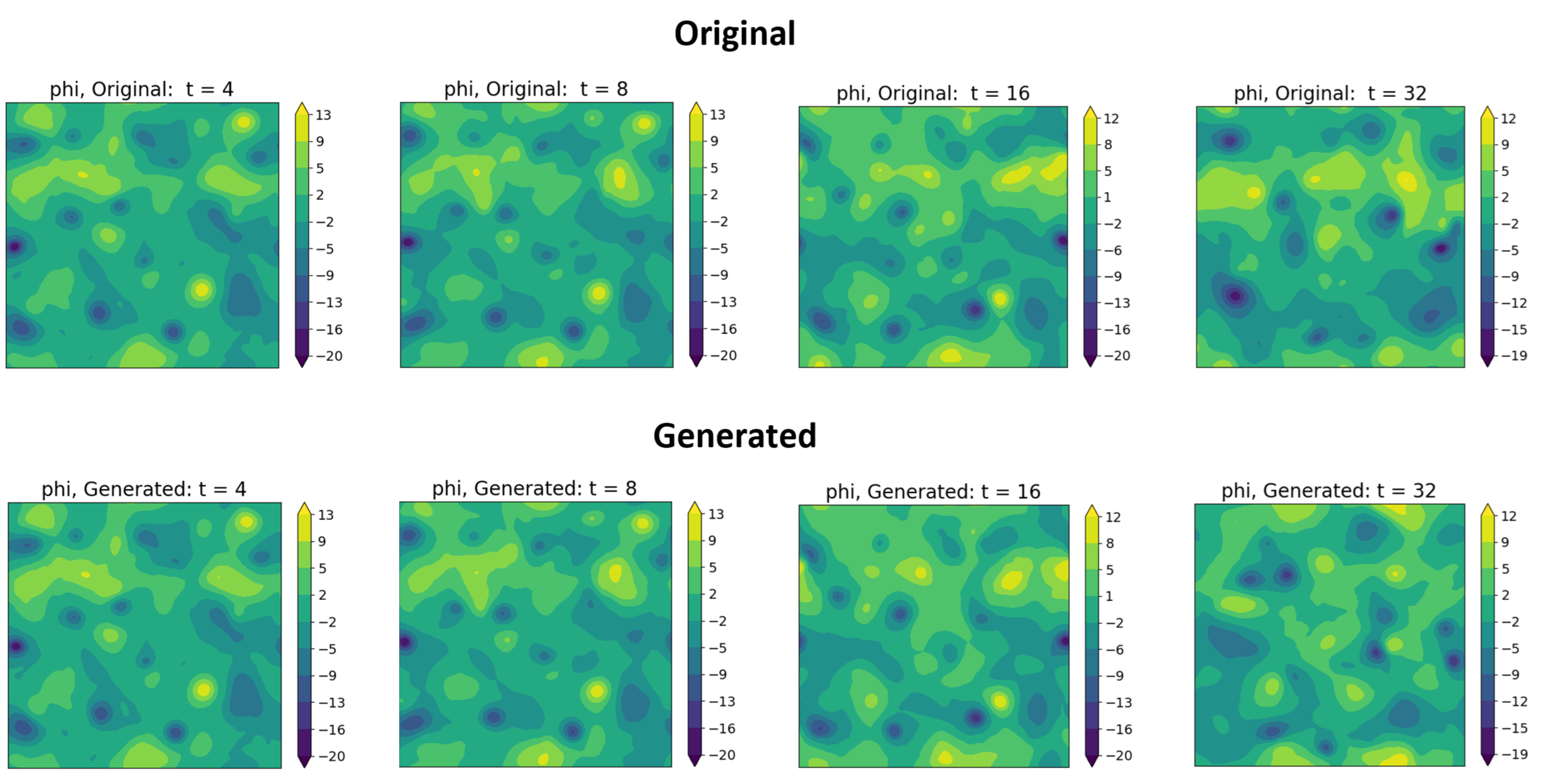}
  \caption{Beginning of a 2000 time step trajectory for HW turbulence generated by the DW-Net-3 model (best realization). The field of $phi$: Generated data - bottom row vs. numerical simulation (using BOUT++) - top row. At first, the generated solution resembles the original (simulated) one quite closely. However, with time, the differences amplify and by the time step 32 become significant.}
  \label{HW-phi}
\end{figure}

\newpage

\bibliographystyle{ieeetr}
\bibliography{lit}

\end{document}